\documentclass[pdflatex,sn-mathphys-num,iicol]{sn-jnl}

\usepackage{graphicx}%
\usepackage{multirow}%
\usepackage{amsmath,amssymb,amsfonts}%
\usepackage{amsthm}%
\usepackage{mathrsfs}%
\usepackage[title]{appendix}%
\usepackage{xcolor}%
\usepackage{textcomp}%
\usepackage{manyfoot}%
\usepackage{booktabs}%
\usepackage{algorithm}%
\usepackage{algorithmicx}%
\usepackage{algpseudocode}%
\usepackage{tabularx}
\usepackage{listings}%
\usepackage{float}
\usepackage{tikz}
\usepackage{ulem}
\usepackage{amsmath} 
\usepackage{ltablex}\keepXColumns
\usepackage{booktabs}
\usepackage{makecell}
\usepackage{graphicx} 
\usetikzlibrary{positioning, shapes, arrows.meta,calc}
\usetikzlibrary{decorations.pathreplacing}

\theoremstyle{thmstyleone}%
\theoremstyle{thmstyletwo}%

\theoremstyle{thmstylethree}%

\begin{document}

\newcommand{\gowm}[1]{\textcolor{blue}{[\textbf{GOWREESH}: #1]}}
\newcommand{\pascal}[1]{\textcolor{orange}{[\textbf{Pascal}: #1]}}
\newcommand{\stevan}[1]{\textcolor{magenta}{[\textbf{Stevan}: #1]}}

\newcommand{\blue}[1]{\textcolor{blue}{#1}}
\newcommand{\red}[1]{\textcolor{red}{#1}}

\title[Article Title]{Looking Beyond the Obvious: A Survey on Abstract Concept Recognition for Video Understanding}



\author[1]{\fnm{Gowreesh} \sur{Mago}}\email{g.mago@uva.nl}

\author[1]{\fnm{Pascal} \sur{Mettes}}\email{p.s.m.mettes@uva.nl}

\author[1]{\fnm{Stevan} \sur{Rudinac}}\email{s.rudinac@uva.nl}

\affil[1]{\orgname{University of Amsterdam}, \country{The Netherlands}}



\abstract{The automatic understanding of video content is advancing rapidly. Empowered by deeper neural networks and large datasets, machines are increasingly capable of understanding what is concretely visible in video frames, whether it be objects, actions, events, or scenes. In comparison, humans retain a unique ability to also look beyond concrete entities and recognize abstract concepts like justice, freedom, and togetherness. Abstract concept recognition forms a crucial open challenge in video understanding, where reasoning on multiple semantic levels based on contextual information is key. In this paper, we argue that the recent advances in Foundation Models make for an ideal setting to address abstract concept understanding in videos.
Automated understanding of high-level abstract concepts is imperative as it enables models to be more aligned with human reasoning and values. In this survey, we study different tasks and datasets used to understand abstract concepts in video content. We observe that, periodically and over a long period, researchers have attempted to solve these tasks, making the best use of the tools available at their disposal. We advocate that drawing on decades of community experience will help us shed light on this important open grand challenge and avoid ``re-inventing the wheel'' as we start revisiting it in the era of multi-modal Foundation Models.}
\maketitle

\section{Introduction}
\begin{figure*}
    \centering
    \includegraphics[width=\textwidth]{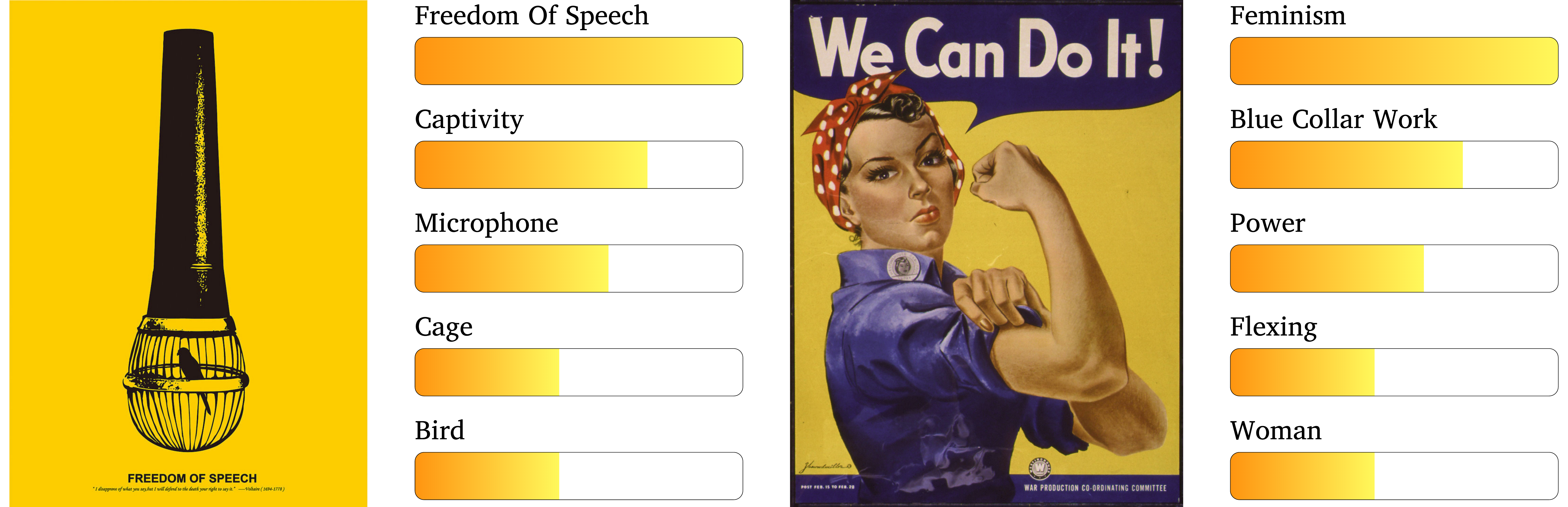}
    \caption{Posters from \cite{reddotDesignAward} and \cite{war_production_1942}. The concepts, alongside their level of abstraction, are represented on a horizontal scale. We unlock abstract concepts as we understand the visual content with increasing semantic depth.}
    \label{fig:abstract_concepts}
\end{figure*}

\begin{figure*}[h]
    \centering
    \includegraphics[width=\textwidth]{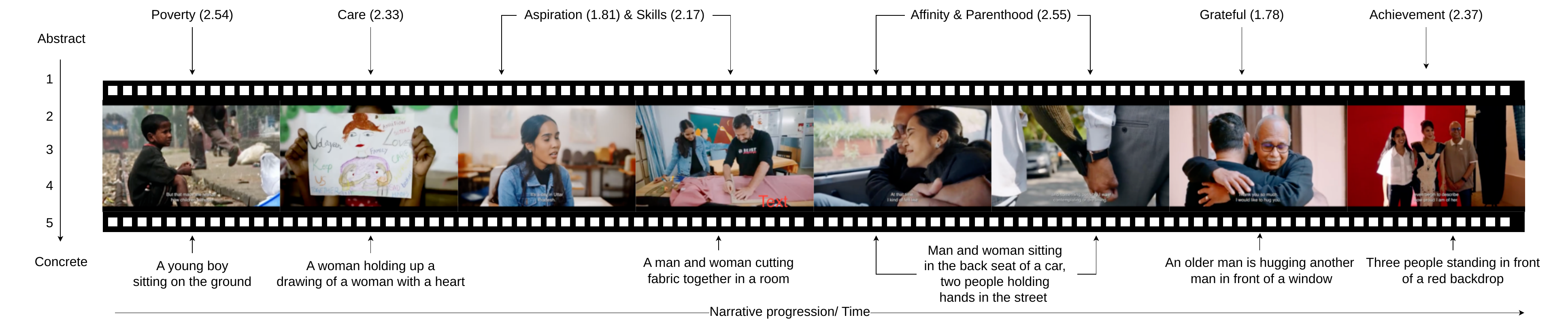}
    \caption{Keyframes from a philanthropic video \cite{beastphilanthropy_2024_empowering} where a woman who grew up in an orphanage is helped break into fashion design. Frame-level topic descriptions with their concreteness score (1-5), as defined by \cite{Brysbaert2014ConcretenessRF}, are presented at the top, while the descriptions generated by BLIP-2 \cite{Li2023BLIP2BL} are provided at the bottom. Notice how concrete descriptions of keyframes corresponding to the outputs from BLIP-2  miss the abstract topic-level descriptions.}
    \label{fig:philanthropy}
\end{figure*}

Abstract concepts encapsulate more than the objective and tangible.  While abstract concepts are easily understood and felt by humans, developing computational models capable of reasoning at such a high semantic level poses fundamental challenges. Figure~\ref{fig:abstract_concepts} illustrates the levels of abstractions observed in the visual content. Behind the directly observable objects are semantic topics and visual metaphors, conveying a deeper meaning that may be hidden and non-obvious.
Videos provide a unique visual medium for abstract concept understanding, as many abstract concepts have a temporal component and appear only over time (cf.
Figure~\ref{fig:philanthropy}). Themes and concepts like relationships between characters and the intent of actions only become clear when watching the entire video and analyzing all modalities.

The state-of-the-art in video understanding is rapidly advancing in recognizing and localizing concrete concepts in videos, from objects \cite{Liu2018DeepLF} to scenes \cite{zeng2021deep} and actions \cite{Kong2018HumanAR}. Models for concrete concept recognition thrive on learning from examples. In contrast, abstract concepts require context \cite{Zhao2021ComputationalEA, Stefanini2019ArtpediaAN} and broader knowledge from hybrid AI systems \cite{pandiani2023seeing}, which makes recognition far more challenging.

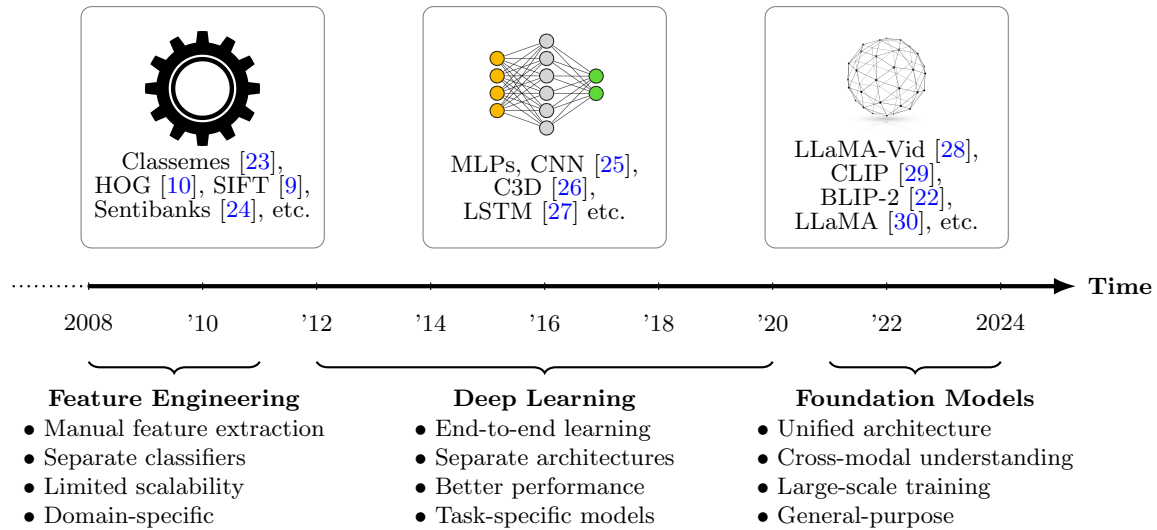
\begin{figure*}[h]
    \centering
    \begin{tikzpicture}[every node/.style={font=\small}, >=latex]

\tikzset{
  stage/.style={
  rectangle,
  rounded corners,
  draw=gray,
  font=\scriptsize,
  minimum width=3.2cm,
  minimum height=3.2cm,
  text width=3cm,  
  align=center,
  text centered,
  inner sep=0pt
},
}

\def\StartYear{2008}
\def\EndYear{2024}
\def\FixedLength{12} 
\def\mark{0.05}
\pgfmathtruncatemacro{\YearDiff}{\EndYear - \StartYear}
\pgfmathsetmacro{\UnitScale}{\FixedLength/\YearDiff}

\draw[dotted, thick] (-1.0,0) -- (0,0);
\draw[ultra thick, -{Latex[length=3mm]}] (0,0) -- (\FixedLength+1,0) node[right] {\textbf{Time}};

\foreach \i in {0,2,...,\YearDiff} {
    \pgfmathsetmacro{\XPos}{\i * \UnitScale}
    \pgfmathtruncatemacro{\Year}{\StartYear + \i}
    \ifnum\i=0
        \draw (\XPos,-\mark) -- (\XPos,\mark) node[anchor=north] at (\XPos,-0.25) {\Year};
    \else
        \ifnum\i=\YearDiff
            \draw (\XPos,-\mark) -- (\XPos,\mark) node[anchor=north] at (\XPos,-0.25) {\Year};
        \else
            \pgfmathtruncatemacro{\ShortYear}{\Year - 100 * int(\Year / 100)}
            \edef\YearLabel{\ifnum\ShortYear<10 0\number\ShortYear\else\number\ShortYear\fi}
            \draw (\XPos,-\mark) -- (\XPos,\mark) node[anchor=north] at (\XPos,-0.25) {'\YearLabel};
        \fi
    \fi
}


\pgfmathsetmacro{\XOne}{(2010-\StartYear)*\UnitScale}
\pgfmathsetmacro{\XTwo}{(2016-\StartYear)*\UnitScale}
\pgfmathsetmacro{\XThree}{(2022-\StartYear)*\UnitScale}

\node[stage,anchor=south] at (\XOne, 0.5) {
  \includegraphics[height=1.5cm]{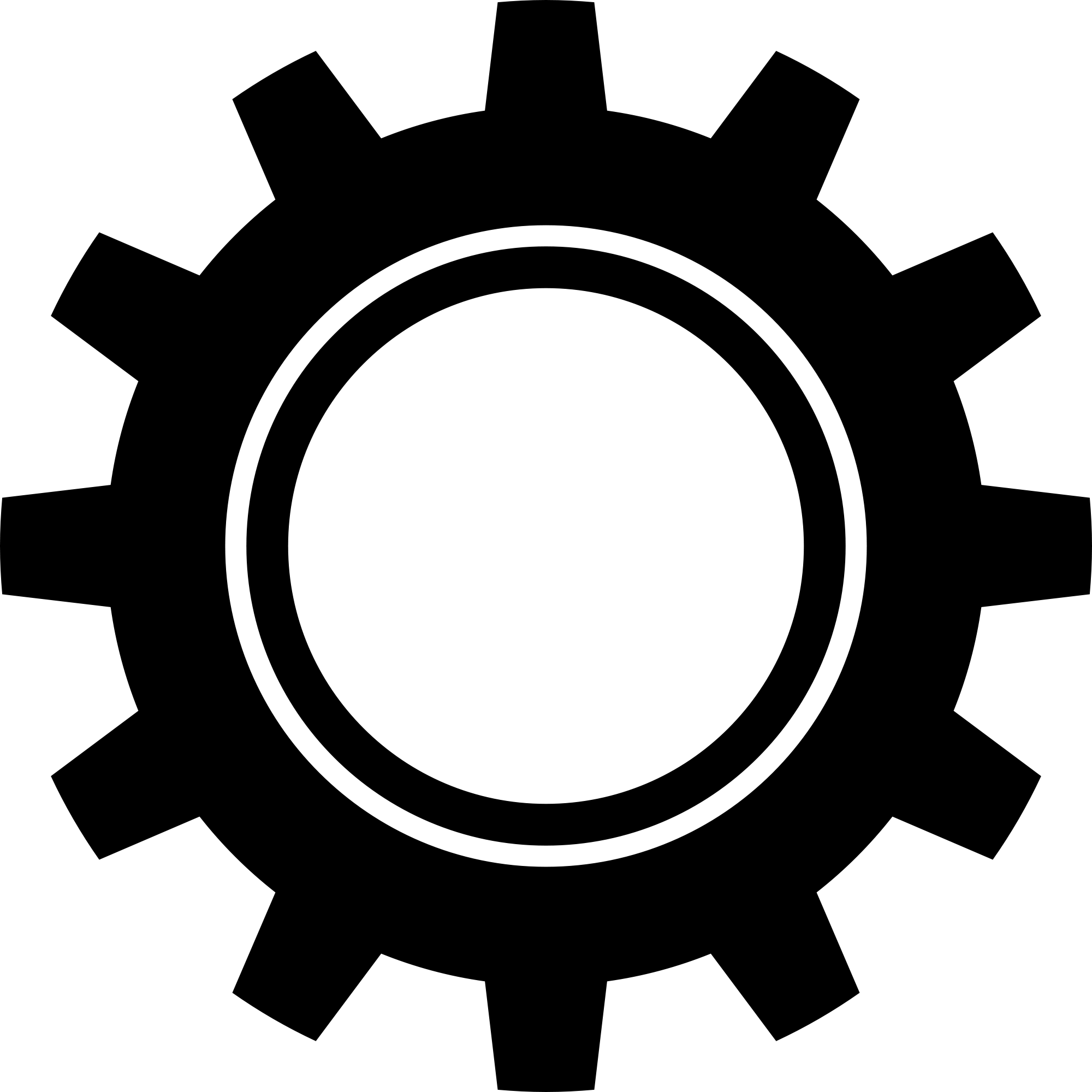} \\
    Classemes \cite{torresani2010efficient}, HOG \cite{Dalal2005HistogramsOO}, SIFT \cite{Lowe2004DistinctiveIF}, Sentibanks \cite{Borth2013SentiBankLO} etc.
};
\node[stage,anchor=south] at (\XTwo, 0.5) {
  \includegraphics[height=1.5cm]{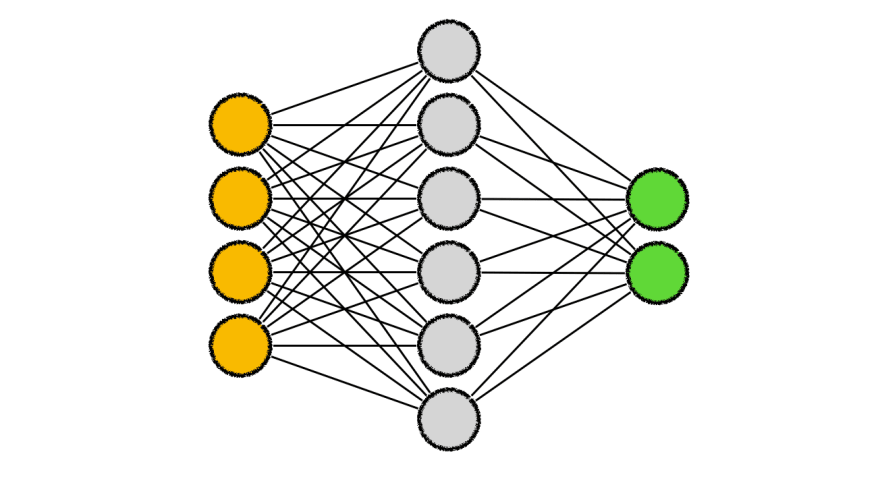} \\[2pt]
    MLPs, CNN \cite{simonyan2014very}, \\C3D \cite{tran2015learning}, LSTM \cite{hochreiter1997long} etc.
};
\node[stage,anchor=south] at (\XThree, 0.5) {
  \includegraphics[height=1.5cm]{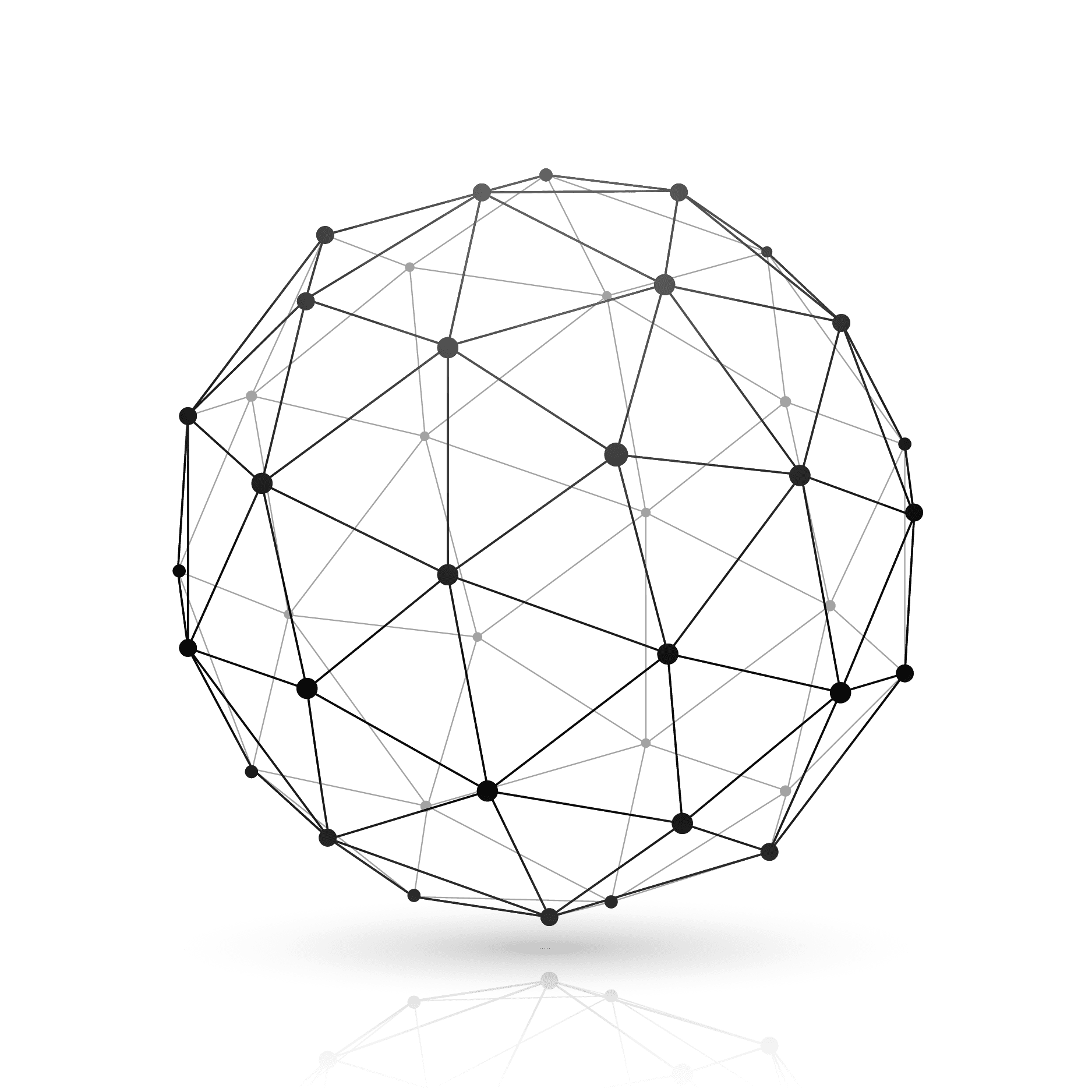} \\
   LLaMA-Vid~\cite{Li2023LLaMAVIDAI}, CLIP \cite{Radford2021LearningTV},
   BLIP-2 \cite{Li2023BLIP2BL}, LLaMA~\cite{Dubey2024TheL3} etc.
};

\newcommand{\addera}[3]{ 
    \pgfmathtruncatemacro{\SYear}{#1}
    \pgfmathtruncatemacro{\EYear}{#2}
    \pgfmathsetmacro{\XStart}{(\SYear - \StartYear)*\UnitScale}
    \pgfmathsetmacro{\XEnd}{(\EYear - \StartYear)*\UnitScale}
    \pgfmathsetmacro{\XMid}{(\XStart+\XEnd)/2}

    \draw[thick, decorate, decoration={brace, amplitude=0.15cm, mirror}]
        (\XStart,-1) -- (\XEnd,-1);

    \node[anchor=north] at (\XMid,-1.2) {#3};
}

\addera{2008}{2011.9}{
  \begin{tabular}{c}
    \textbf{Feature Engineering} \\
    \begin{tabular}{@{}l@{}}
      • Manual feature extraction \\
      • Separate classifiers \\
      • Limited scalability \\
      • Domain-specific \\
    \end{tabular}
  \end{tabular}
}

\addera{2012.1}{2020.9}{
  \begin{tabular}{c}
  \textbf{Deep Learning} \\
    \begin{tabular}{@{}l@{}}
      • End-to-end learning \\
      • Separate architectures \\
      • Better performance \\
      • Task-specific models \\
    \end{tabular}
  \end{tabular}
}

\addera{2021.1}{2024}{
\begin{tabular}{c}
    \textbf{Foundation Models} \\
    \begin{tabular}{@{}l@{}}
      • Unified architecture \\
      • Cross-modal understanding \\
      • Large-scale training \\
      • General-purpose \\
    \end{tabular}
  \end{tabular}
}

\end{tikzpicture}
 
    \caption{Research on abstract concept understanding through three key computer vision eras: feature engineering, deep learning, and Foundation Models. While earlier approaches used separate models to bring external context, this limitation is overcome by Foundation Models due to large-scale training and cross-modal design.
}
    \label{fig:generations}
\end{figure*}

Recognizing abstract concepts is, however, not a new endeavor. Many researchers have previously addressed this problem, which is typically fueled by new advances in feature extraction and multi-modal learning for videos, as highlighted in Figure \ref{fig:generations}. Attempts have been made to understand the occurrence of such concepts in video datasets where the difficulty arises because of subjectivity and lack of context. Over time, the methods have transitioned from hand-crafted features like SIFT \cite{Lowe2004DistinctiveIF} and HOG \cite{Dalal2005HistogramsOO} to model emotions \cite{Jiang2014PredictingEI} for example, to deep learning features from CNNs \cite{Lu2014RAPIDRP} and recently to Foundation models \cite{Zhang2024QualityAI, Cheng2024EmotionLLaMAME} for better context. This evolution, we emphasize, is a recurring theme in the following sections of the paper.

Foundation models \cite{Zhang2023VisionLanguageMF} form the latest advances in the field. Rather than training models for each task specifically, big networks are trained on broad data to adapt to many tasks. Such models provide the context and broad knowledge crucial for tackling abstract concept understanding in videos. This becomes crucial for recognizing abstract concepts and performing other high-level semantic tasks as different visual concepts \cite{MartinezPandiani2023HypericonsFI} and situations \cite{Cerini2024RepresentingAC} can highlight the same abstract concept and the same visual concept can highlight multiple abstract concepts \cite{Zhao2021ComputationalEA}.

Understanding abstract concepts enables us to bring computers close to human-level perception, unlocking various possibilities. Much more complex and subjective queries can be addressed, increasing the search relevance of text to video retrieval-based systems. The contextual understanding and real-world grounding of Foundation models may enable us to design video content analysis engines for monitoring at a large scale. This has applications such as predicting a video's viral potential based on the content, determining the likely user reactions, analyzing political campaigns and leaders’ leanings, and curbing misuse of algorithms for maximum reach. However, this is not a unimodal challenge. The concept of the semantic gap \cite{Smeulders2000ContentBasedIR} is a recurring theme in multimodal research, for example, in visual \cite{Aditya2019IntegratingKA} and audio \cite{Fu2011ASO} modalities. Hence, we must identify high levels of abstractions from all modalities to address the challenges effectively.

This survey explores abstract concepts and tasks formulated at high semantic levels that go beyond objective reasoning, specifically focusing on concepts in visual data, with other modalities in videos as support. We shall also highlight some important image datasets that tackle understanding abstract concepts in the image domain. To provide an overview and exhaustive taxonomy of current literature, we have performed automatic and manual literature analyses. We discuss several generations of papers that tackle abstract concept recognition, from hand-crafted features and end-to-end learning with deep neural networks to the latest trends involving Foundation models. We additionally provide an overview of the datasets available for researching abstract concept understanding in videos, and we provide guidelines on how the next generation of models can learn from the rich past of the field. The emergence of new datasets and benchmarks, combined with the increasing availability of computational resources and advancements in model capabilities, represents the next frontier in the video understanding domain. This survey addresses a critical gap in the existing literature.
 Overall, the paper has the following contributions:

\begin{enumerate}
    \item We present a taxonomy of abstract concepts and high-level semantic tasks for videos that have been dealt with in the computer vision and related communities. This taxonomy seeks to consolidate previously fragmented research directions under a comprehensive framework of abstract concepts.
    \item We discuss in detail the seminal literature that has contributed significantly to the development of methods for understanding such concepts from videos and also discuss key recurrent challenges faced by them.
    \item We propose future research directions and describe how Foundation models can help bridge the existing gap in understanding with the large contextual knowledge.
\end{enumerate}

\section{Defining Abstract Concepts, High-Level Semantics and their Intersection}
Abstract concepts have long been studied, and research on them can be traced to neuroscience \cite{Mkrtychian2019ConcreteVA} and literature dealing with semiotics \cite{Banks2023ConsensusPC} and rhetorical analysis \cite{Bateman2014TextAI}. While concrete concepts are usually easy to associate with materials, abstract concepts are often associated with notions and sentiments that can be felt but are challenging to describe in concrete terms \cite{Brysbaert2014ConcretenessRF}. Such ideas are difficult to understand due to the inherent subjectivity \cite{Wiemerhastings2005ContentDF} in interpreting content and the lack of contextual understanding \cite{LanglandHassan2022ACA} capabilities in automatic understanding systems. This point is also emphasized in \cite{MartinezPandiani2024TheWP}, where violence is shown to be interpretable through vastly different visual concepts, such as police brutality and Renaissance painting. Therefore, this survey emphasizes recognizing these abstract concepts that inherently need context and whose identification is affected by subjectivity. This forms the core idea behind the selection of clusters for studying issues related to abstract concepts. 

Semantics and the semantic gap are at the core of multimodal content, as highlighted in \cite{Smeulders2000ContentBasedIR, Hare2006MindTG}, and bridging the semantic gap has been one of the key challenges in visual and multimodal content analysis. A central idea of semantics is the three-tiered representation in, e.g., vision \cite{Aditya2019IntegratingKA} and audio \cite{Fu2011ASO} modalities. The lowest tier focuses on low-level descriptors like shapes and edges for vision and audio frequency for acoustic features. Due to the vast work on these signals, computational understanding of such signals is almost trivial. As we reach higher levels of semantics, we encounter different structures like objects, scenes, actions, rhythm, and pitch. The highest level of semantics includes dimensions close to human-level perception signals like mood, genre, and artistic style. The semantic gap, therefore, aims to bring the model capabilities to higher semantic levels or closer to human-level perception and understanding. Sharma et al. \cite{Sharma2024FromPT} provides an overview of abstract concepts and semantics in image understanding with applications in marketing. We significantly broaden and deepen prior work by generalizing it to the domain of video understanding.

Literature in communication provides some insight into multimodal content and the occurrence of such concepts. Lasswell \cite{lasswell1960structure, Singh2024TeachingHB} encodes communication with the following phrase: \textit{``Who says what to whom with what effect.''}. The occurrence of subtle abstract concepts can, therefore, happen at the origin, such that these are carried as a signal in the content itself or can be at the receiver, where these concepts manifest as emotions or perceptions. Capturing subtle signals from all modalities is vital to correctly understanding such concepts. Bateman \cite{Bateman2014TextAI} introduces the idea of meaning multiplication, which suggests that modalities might, in some scenarios, carry little significance or even contrary meanings when isolated. Still, together, they unveil something more descriptive. We conjecture that achieving a human-level understanding of content requires models to possess social intelligence and common-sense reasoning, which is also supported by previous research \cite{Banks2023ConsensusPC}.

\section{Survey Methodology}
To cast a wide net and survey the topic from many different perspectives, we focus on a set of conferences we believe have an exhaustive video and image research community covering all aspects of computer vision tasks. Further sections discuss the automated literature survey pipeline.

\subsection{Pipeline}
\begin{figure}[t]
    \centering
    \includegraphics[width=0.45\textwidth]{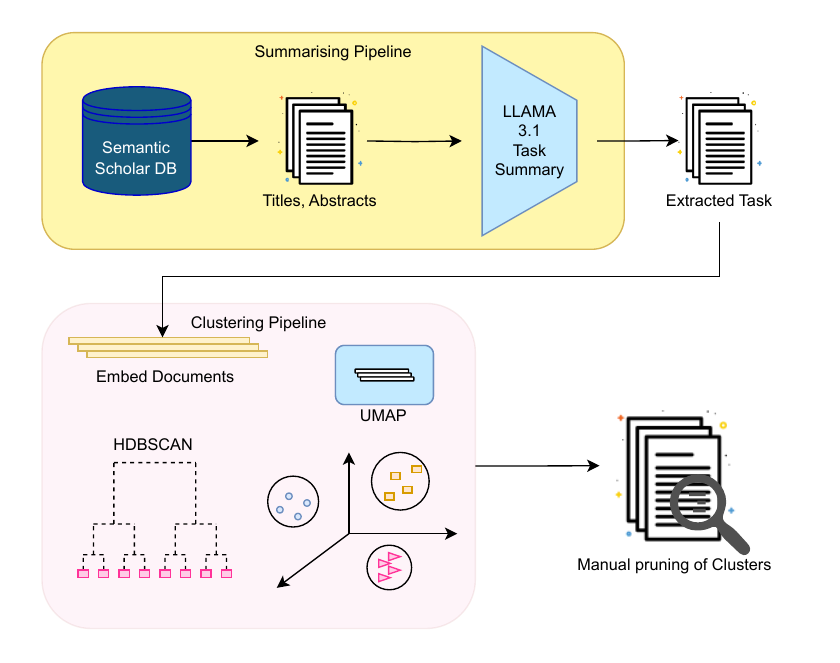}
    \caption{Our survey pipeline consists of automatic article summarization, clustering, and manual filtering.}
    \label{fig:survey_method}
\end{figure}

The literature on computer vision encompasses various domains, including detection, segmentation, and generation. As explained in this section, filtering out literature focusing specifically on abstract concepts requires developing a detailed methodology resulting in this pipeline. Figure \ref{fig:survey_method} highlights the automatic and manual steps for the literature survey pipeline. We use the data indexed until September 2024 from Semantic Scholar \cite{Kinney2023TheSS} as it contains a large corpus of papers from various $A^*$, $A$, and $B$ rated conferences according to \cite{coreCoreeduauCORE}. Semantic Scholar, besides offering information about the venue, authors, and the title of the manuscript, also includes abstracts that aid in the automatic filtering process. We filter out papers that do not have the word `video' in the title or abstract, which leaves us with $\approx$16K papers. We then generate short summaries of the title and abstract, focusing on the task being tackled in the paper using quantized LLAMA 3.1 \cite{Dubey2024TheL3}, 
removing details like the model architecture and dataset. This is done so that the papers are clustered based on the topics they deal with and not based on models or training paradigms. We use BERTopic \cite{grootendorst2022bertopic} as the framework for topic modeling and clustering. The documents are embedded using the sentence embedding model from \cite{zhang2024mgte}. This model was chosen due to its high performance on the MTEB leaderboard \cite{muennighoff2022mteb} and its relatively low number of parameters (434M), which allows for faster inference. We then manually eliminate clusters focused on conventional problems in video, including object detection, 3D generation, and action recognition, leaving us with $\approx$ 1.8K papers. After that, a manual search is conducted, filtering papers based on the concepts of interest for this survey. We do so by looking at the datasets the authors use and whether they address abstract concepts or if the task requires a deeper understanding of the video content. We also check for newly published literature that cites these papers to make the search exhaustive.

\subsection{Survey Organisation}
Recently, an effort has been made to organize abstract concepts in the visual domain by Pandiani and Presutti \cite{pandiani2023seeing}. Their work focuses solely on image-based classification, organizing abstract concepts into four categories: commonsense, emotional, aesthetic, and inductive-interpretative semantics. While the overview of domains of abstract concepts overlaps with our survey, the organization and outlook differ. Also, their scope is limited to CNN-based models, only briefly touches on transformers, and excludes Foundation models. We updated this taxonomy for videos, structuring them under three pillars: Narrative \& Rhetoric, Perception Understanding, and Emotions \& Social Signals. These help capture spatiotemporal and multimodal reasoning that image surveys lack. For example, image metaphors might combine unusual objects, such as a cheetah merged with a racecar, while video metaphors develop over time, as seen in a sequence where a racecar and a cheetah race side by side. Our survey extends previous surveys on video understanding \cite{nguyen2024video, sanders2024survey, madan2024foundation, stergiou2025time}, which focus on algorithmic advances and recognizing concrete concepts like events and actions. Our survey, however, focuses on abstract concepts and high-level semantics, discussing the datasets, algorithms, and unique challenges posed by them. We primarily investigate videos, but also examine some image-based datasets where the history is richer. We also do not restrict our study based on the prediction formats, and we discuss datasets and benchmarks in retrieval, captioning, question-answering, and other areas. Overall, our work is the first comprehensive survey extending abstract concept recognition to videos and reflecting on the long and much richer than commonly assumed history of the field, stretching from early works on low-level visual features, to advances brought to the table by deep learning and state-of-the-art multimodal Foundation models.

Research in communication science deals with meaning construction from raw signals \cite{van2018communication}. Drawing inspiration from communication science \cite{van2018communication, Craig1999CommunicationTA}, psychology \cite{borghi2017challenge}, cognitive science \cite{paivio1965abstractness}, and neuroscience \cite{mkrtychian2019concrete}, which have long studied abstract concepts, our framework is organized around three foundational pillars: Perception understanding, Emotions \& Social Signals, and Narrative \& Rhetoric Analysis. These fields provide the functional architecture that AI systems need to compute to understand abstract concepts. Without such reasoning ingrained, modern AI systems will remain trapped in literal interpretation: they can recognize ``crowd gathering" but not whether it represents protest, celebration, or an emergency, which requires understanding intent, socio-cultural dynamics, emotions etc. Borghi et al. propose that understanding abstract concepts requires grounding in perception, action, language, and culture \cite{borghi2022abstract}.

\definecolor{imagecolor}{HTML}{00008B} 
\definecolor{videocolor}{HTML}{fe218b} 

\newcommand{\videofmt}[1]{\textbf{#1}}
\newcommand{\imagefmt}[1]{\textit{#1}}

\clearpage
\onecolumn
\begin{table*}[t]
\centering
\scriptsize
\caption{Datasets and benchmarks organized in abstract concept recognition subdomains and formatted as (\videofmt{video}, \imagefmt{image}). Each dataset is further categorized by its evaluation protocol, as indicated by the dataset type legend: (C) Classification, where models associate inputs to discrete abstract categories (e.g., predicting emotion type); (Q) Question Answering, where given a query related to abstract concept in natural language the machine returns a response in natural language; (R) Regression, involving prediction of continuous scores for abstract variables (e.g., aesthetic intensity, valence-arousal); (Cap) Captioning, where models generate text given an input video (such as generating metaphors given an input video); (Ret) Retrieval, assessing how well models align visual and textual abstractions for cross-modal search; and (Ra) Ranking, where systems rank samples based on the relative strength of an abstract attribute (e.g., degree of persuasiveness).}

\begin{tabularx}{\textwidth}{l l X}
\toprule
\textbf{Category} & \textbf{Subcategory} & \textbf{Datasets/Benchmarks} \\
\midrule

\multirow{4}{*}{\shortstack[l]{Perception\\Understanding}} 
& Intent & \videofmt{Oops (C) \cite{Epstein2019OopsPU}}, \videofmt{IntentQA (Q) \cite{Li2023IntentQACV}}, \videofmt{FunQA (Q) \cite{Xie2023FunQATS}}, \videofmt{Vid2Int (C) \cite{Xu2021Vid2IntDI}}, \videofmt{MIntRec2.0 (C) \cite{Zhang2024MIntRec20AL}}, \videofmt{BlackSwan (Q) \cite{chinchure2025black}}, \videofmt{MINE (C) \cite{yang2025uncertain}}, \imagefmt{VCR (Q) \cite{zellers2019recognition}}, \imagefmt{Intentonomy (C) \cite{Jia2020IntentonomyAD}}\\

\cmidrule(l){2-3}
& Visual Aesthetics &  \videofmt{KoNViD-1k (R) \cite{hosu2017konstanz}}, \videofmt{LSVQ (R) \cite{ying2021patch}}, \videofmt{DIVIDE-3k (R) \cite{wu2023exploring}}, \videofmt{Q-Bench-Video (Q) \cite{zhang2025q}}, \imagefmt{AVA (C) \cite{Murray2012AVAAL}},  \imagefmt{LIVE-itW (R) \cite{ghadiyaram2015massive}}, \imagefmt{MDID (C) \cite{sun2017mdid}}, \imagefmt{KonIQ-10k (R) \cite{lin2018koniq}}, \imagefmt{SPAQ (R) \cite{fang2020perceptual}}, \imagefmt{LAMBDA (Q) \cite{HariniS2023LongTermAM}} \\

\cmidrule(l){2-3}
& Semantic Theme Understanding & \videofmt{Pitts Ads Dataset (C) \cite{Hussain_2017_CVPR}}, \videofmt{YouTube-8M (C) \cite{abu2016youtube}}, \videofmt{LVU (C) \cite{Wu2021TowardsLV}}, \videofmt{Tencent AVS (C) \cite{Wang2021OverviewOT}}, \videofmt{MM-AU (C) \cite{Bose2023MMAUTowardsMU}}, \videofmt{VideoAds \cite{zhang2025videoads}}, \videofmt{3MASSIV (C) \cite{Gupta20223MASSIVMM}}, \videofmt{AdsQA (Q) \cite{long2025adsqa}}, \imagefmt{LAION-400M (Ret) \cite{schuhmann2021laion}}, \imagefmt{DEEPEVAL (Q) \cite{Yang2024CanLM}}, \imagefmt{Artpedia (C) \cite{Stefanini2019ArtpediaAN}}, \imagefmt{SemArt (Ret) \cite{Garcia_2018_ECCV_Workshops}}, \imagefmt{UIV (C) \cite{AlamedaPineda2017ViraliencyPL}} \\

\cmidrule(l){2-3}
& User Behaviour Modeling/Virality & \videofmt{LVU (R) \cite{Wu2021TowardsLV}}, \videofmt{MicroVideos (R) \cite{Chen2016MicroTM}}, \videofmt{CMU Viral Video (R) \cite{Jiang2014ViralVS}}, \videofmt{Top\&Random (R) \cite{figueiredo2011tube}} , \videofmt{GPVD (C) \cite{kayal-etal-2025-large}}, \imagefmt{Reddit Images (R) \cite{lakkaraju2013s}}, \imagefmt{UIV (C) \cite{AlamedaPineda2017ViraliencyPL}} \\
\midrule

\multirow{3}{*}{\shortstack[l]{Emotions and\\Social Signals}} 
& Affective Analysis & \videofmt{Pitts Ads Dataset (C) \cite{Hussain_2017_CVPR}}, \videofmt{Video Emotion Dataset (C) \cite{Jiang2014PredictingEI}}, \videofmt{Ekman Emotion Dataset (C) \cite{Jiang2014PredictingEI}}, \videofmt{VAAD (C) \cite{Pang2020FurtherUV}}, \videofmt{iMiGUE (C) \cite{Liu2021iMiGUEAI}}, \videofmt{EALD (Q) \cite{Li2024EALDMLLMEA}}, \videofmt{VCE (C) \cite{Mazeika2022HowWT}}, \videofmt{V2V (R) \cite{Mazeika2022HowWT}}, \videofmt{VEATIC (R) \cite{Ren2023VEATICVE}}, \videofmt{MERR (C,Cap) \cite{Cheng2024EmotionLLaMAME}}, \videofmt{3MASSIV (C) \cite{Gupta20223MASSIVMM}}, \videofmt{EMER (C) \cite{lian2025affectgpt}}, \imagefmt{LAMBDA (Q) \cite{HariniS2023LongTermAM}}, \imagefmt{ArtEmis (C,Cap) \cite{Achlioptas_2021_CVPR}}, \imagefmt{EmoSet (C) \cite{Yang2023EmoSetAL}} \\

\cmidrule(l){2-3}
& Relationships & \videofmt{SRIV (C) \cite{Lv2018MultistreamFM}}, \videofmt{ViSR (C) \cite{Liu2019SocialRR}}, \videofmt{PERR (C) \cite{gao2021pairwise}}, \videofmt{MovieGraphs (Q) \cite{Vicol2017MovieGraphsTU}}, \videofmt{LVU (C) \cite{Wu2021TowardsLV}}, \videofmt{VideoAds \cite{zhang2025videoads}}, \imagefmt{Social Relation Dataset (C) \cite{Zhang2015LearningSR}}, \imagefmt{PISC (C) \cite{Li2017DualGlanceMF}}, \imagefmt{PIPA (C) \cite{Sun2017ADB}} \\

\cmidrule(l){2-3}
& Situation Analysis & \videofmt{MovieGraphs (Q) \cite{Vicol2017MovieGraphsTU}}, \videofmt{HLVU (Q) \cite{Curtis2020HLVUAN}}, \videofmt{Social-IQ (Q) \cite{Zadeh2019SocialIQAQ}}, \videofmt{DeSIQ (Q) \cite{Guo2023DeSIQTA}} \\
\midrule

\multirow{6}{*}{\shortstack[l]{Narrative \&\\Rhetorical Analysis}} 
& Humor/Sarcasm/Satire &  \videofmt{MUStARD (C) \cite{Castro2019TowardsMS}}, \videofmt{UR-FUNNY (C) \cite{Hasan2019URFUNNYAM}}, \videofmt{MHD (C) \cite{Patro2021MultimodalHD}}, \videofmt{WITS (C) \cite{Kumar2022WhenDY}}, \videofmt{ExFunTube (Cap) \cite{Ko2023CanLM}}, \imagefmt{YesBut (C) \cite{Nandy2024YesButAH}}, \imagefmt{V-FLUTE (Cap,C) \cite{Saakyan2024UnderstandingFM}},\imagefmt{AVH (R) \cite{Chandrasekaran2015WeAH}}, \imagefmt{FOR (C) \cite{Chandrasekaran2015WeAH}} \\

\cmidrule(l){2-3}
& Visual Metaphors & \videofmt{VMC (Cap) \cite{kalarani2024unveiling}}, \imagefmt{V-FLUTE (Cap,C) \cite{Saakyan2024UnderstandingFM}}, \imagefmt{MultiMET (C) \cite{Zhang2021MultiMETAM}}, \imagefmt{MetaCLUE (C,Q,Cap) \cite{Akula2022MetaCLUETC}} \\

\cmidrule(l){2-3}
& Misinformation & \videofmt{VMH (C) \cite{Sung2023NotAF}}, \videofmt{FakeSV (C) \cite{Qi2022FakeSVAM}}, \videofmt{Fake Video Corpus (C) \cite{papadopoulou2019corpus}}, \imagefmt{NewsCLIPpings (C) \cite{luo2021newsclippings}} \\

\cmidrule(l){2-3}
& Polarity/Opinion & \videofmt{MOSI (R) \cite{Zadeh2016MOSIMC}}, \videofmt{POM (C) \cite{Park2014ComputationalAO}}, \imagefmt{Political Advertisements (C) \cite{SnchezVillegas2021AnalyzingOP}}, \imagefmt{Political leaning (C) \cite{Thomas2019PredictingTP}} \\

\cmidrule(l){2-3}
& Persuasion & \videofmt{Rallying a Crowd (RAC) (C) \cite{Siddiquie2015ExploitingMA}}, \videofmt{QPS (R) \cite{Bai2020M2P2MP}}, \videofmt{Paladin (C) \cite{liu2025paladin}} \imagefmt{ImageArg (C) \cite{Liu2022ImageArgAM}}, \imagefmt{Persuasive\_meme (C) \cite{Kumari2023ThePM}},\imagefmt{Pitts Ads Dataset (C) \cite{Singla2022PersuasionSI}}, \imagefmt{Persuasive Portraits of Politicians (Ra) \cite{Joo2014VisualPI}} \\

\cmidrule(l){2-3}
& Visual Narrative & \videofmt{MPII (Cap) \cite{Rohrbach2015ADF}}, \videofmt{MovieBook (Ret) \cite{Zhu2015AligningBA}}, \videofmt{MovieQA (Q) \cite{Tapaswi2015MovieQAUS}}, \videofmt{DramaQA (Q) \cite{Choi2020DramaQACV}}, \videofmt{SF20K (Q) \cite{Ghermi2024LongSS}}, \videofmt{ObyGaze12 (C) \cite{Tores2024VisualOI}}, \videofmt{MovieNet (C) \cite{Huang2020MovieNetAH}}, \videofmt{LVU (C) \cite{Wu2021TowardsLV}}, \videofmt{HLVU (Q) \cite{Curtis2020HLVUAN}}, \imagefmt{Mementos (Cap) \cite{Wang2024MementosAC}} \\

\bottomrule
\end{tabularx}
\label{tab:abstract_datasets}
\end{table*}
\clearpage
\twocolumn
By grounding our taxonomy based on these theories, we ensure that we treat videos as encoding richer semantic structure beyond scenes and actions, and also motivate all that needs to be understood in order to facilitate abstract concepts understanding in general. Additionally, we expand the scope of task design beyond traditional classification by incorporating a variety of benchmarks, including video question answering and retrieval, to encompass the full range of tasks relevant to understanding abstract concepts.

We take a data-first approach where we observe the emergent clusters from the pipeline and remove classical/traditional computer vision tasks (like object detection, super-resolution, etc.). The remaining clusters are checked for either the task they address (which is abstract in itself, such as visual metaphor captioning) or for the classical tasks (question answering/retrieval), dealing with datasets that contain certain abstract notions. We exclude datasets that lack abstract concept annotations. The remaining papers are then organized again with principles from communication science, neuroscience, and psychology, leading us to develop this taxonomy. So in summary, the clusters here are automatically created and then manually curated for video understanding tasks based on the theory from several relevant disciplines.

\begin{figure}[t]
    \centering
    \includegraphics[width=0.5\textwidth]{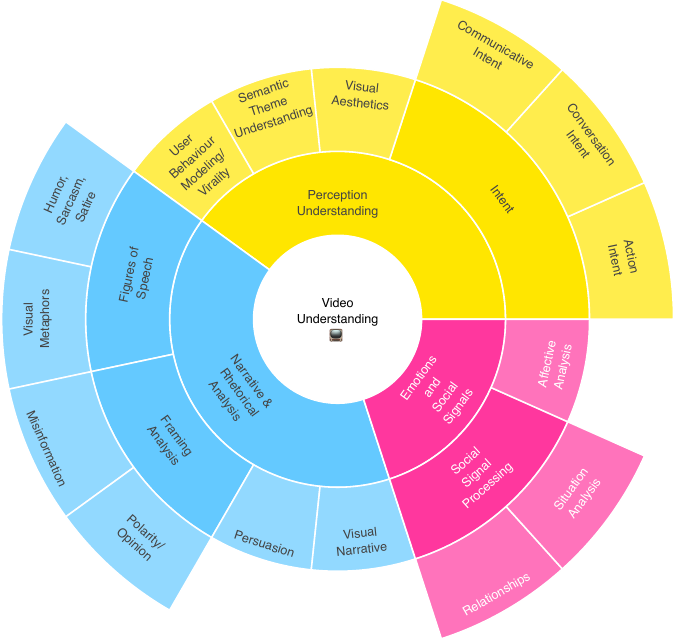}
    \caption{Our proposed taxonomy and broad organization of tasks defining the scope of this survey.}
    \label{fig:topics}
\end{figure}

Based on the automatic literature survey pipeline and going through the clusters to understand tasks previously tackled in video understanding, we present a taxonomy of all the topics of interest for this survey in Figure \ref{fig:topics}. This figure provides a hierarchical organization of all the tasks that tackle abstract concepts and high-level semantic tasks. The leaf nodes represent the exact tasks, while the parent nodes describe the grouping criteria based on where the abstractness/context lies. We consider indirect ways of communication, affective analysis, and understanding of human perception of scenes, actions, and semantic themes. We then tackle each concept/task as we move through the sections of this paper. 

\textit{\textbf{1:} Perception Understanding}: Since aesthetics \cite{chatterjee2014neuroaesthetics} and actions \cite{zdrazilova2018communicating} are vital to perception, we classify aesthetics and intent understanding under our first pillar: Perception Understanding. This pillar also encompasses semantic themes \cite{rudinac2012leveraging} (which can span entire videos as abstract concepts) and user behavior modeling, both of which reflect how users perceive content. This category investigates abstract concepts in human perception of video content. We cover topics such as the semantic theme, aesthetics, and comprehending the intent of \textit{actors} in video content. One intriguing example presented here is based on weak signals, specifically user behavior modeling on social media sites. Comments, exchanges, and indicators such as the like-to-dislike ratio provide insight into the broader perception of video, hence the downstream task of predicting the viral potential of video content.

\textit{\textbf{2:} Emotional and Social Signals}: This pillar covers literature addressing emotional expressions, effects, and social dynamics, such as emotion recognition, interpersonal relationships, and situational awareness within video contexts.

\textit{\textbf{3:} Narrative and Rhetoric Analysis}: This area involves understanding complex communicative intent, including indirect forms of messaging such as metaphor, symbolism, or persuasive narrative techniques. We jointly discuss rhetorical and narrative elements to emphasize the methods used to make the message more appealing. 

To further structure the literature, key benchmarks and datasets are summarized in Table~\ref{tab:abstract_datasets}. These include datasets that have either been proposed as solutions to specific problems or widely adopted for studying particular abstract concepts. Notably, many of these datasets feature annotations for multiple abstract concepts, such as emotion combined with aesthetics or persuasion strategy \cite{Hussain_2017_CVPR}, which could be valuable for developing unified benchmarks.

\section{Perception Understanding}
Perception is closely linked to the processing and interpreting input signals to understand information or the environment \cite{Schacter2011PsychologyE}. Therefore, problems centered on human perception should include how humans perceive certain scenes, situations, and video content. We have identified four main themes in perception understanding, dealing with perception at increasing levels of explicit semantics, from the latent notions of aesthetics to the explicit modeling of user behavior:

\begin{itemize}
    \item \textit{Visual Aesthetics}: Tasks related to human perception of beauty. This is often associated with understanding the semantics of the scene.
    \item \textit{Intent Understanding}: Tasks related to understanding and interpreting the intent behind actions and conversations.
    \item \textit{Semantic Theme Understanding}: Tasks focused on grasping video content's central theme and deeper meaning. The discussed semantics are complex enough that they can't be inferred from a single image alone; instead, AI systems demand a thorough understanding of the video content. 
    
    \item \textit{User Behavior Modeling}: Metadata such as likes, dislikes, comments, and other interactive metrics open doors to understanding the general perception of video content. This is a direct weak signal to how humans perceive video content. We also include an essential problem of understanding video virality as it is a closely tied phenomenon to user metrics.

\end{itemize}

\begin{figure*}
    \centering
    \includegraphics[width=\linewidth]{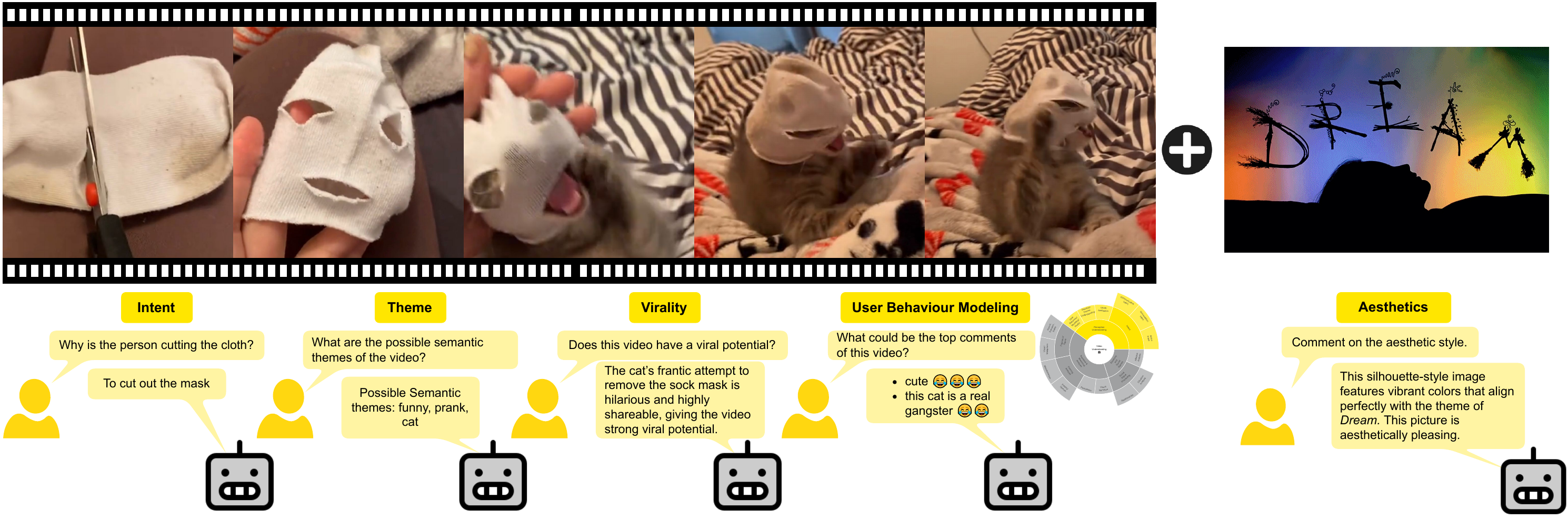}
    \caption{Pillar 1: Perception understanding. We highlight the four key themes in perception understanding, as exemplified by user queries for intent, theme, virality, and user behavior modeling, of a video from the FunQA dataset \cite{Xie2023FunQATS}. Aesthetic understanding capabilities are highlighted via an example from the AVA dataset \cite{Murray2012AVAAL}.}
    \label{fig:pillar1}
\end{figure*}

 Figure \ref{fig:pillar1} gives a high-level idea of what an ideal model with human-like perception would reason like. The case presented involves a human cutting a sock to create a cat mask, which evokes humor and, consequently, is quite popular on social media platforms. Achieving human-like perception would require solving basic challenges, such as understanding actions, scenes, events, and temporal progression, as well as understanding how these combine to form abstract concepts like \textit{humor, surprise, or beauty} at a higher level. Such capabilities would allow models to not only reason about what happens in the video at a surface level but also discover innate concepts like the intent of an action, possible categorization and associating video with relevant topics, predicting the viral potential of the video, the ability to emulate human-like response on watching the video, and also have a grounded perception of aesthetics. We take captioning as an example, but a grounded model should be able to adapt to other tasks, such as retrieval and classification.

\subsection{Visual Aesthetics}
Aesthetics are vital to human perception and require the global structure as a context to infer
ranging from numerical features to natural language descriptions. Since aesthetics are inferred at the scene level, image-level datasets have shown to be important, particularly due to their extensive use for pre-trainined Foundation models. We have therefore included these works in the discussion. We discuss literature and progress made in chronological order. We also pay attention to video memorability, which is closely associated with video aesthetics. 

Early attempts at understanding aesthetics use computational models to mimic understanding of photographic rules like the golden ratio and the rule of thirds \cite{Datta2006StudyingAI, Ke2006TheDO}. Moorthy et al. \cite{Moorthy2010TowardsCM} highlight the need to understand video aesthetics, which was essentially an understudied problem. They propose incorporating motion features (motion ratio, size ratio) besides frame-level features (like colorfulness and color harmony). Murray et al. \cite{Murray2012AVAAL} propose AVA (Aesthetic Visual Analysis) containing 250K images with annotations on the aesthetic scores, photographic style, and image content, marking one of the first large-scale aesthetic quality assessment datasets to push research in the direction. An important insight from exploratory data analysis on the database reveals that non-conventional image styles receive highly variant scores, highlighting the subjectivity associated with aesthetics as a quantity. This dataset remains valuable and is still utilized to test models for their capability in assessing aesthetic quality. In addition, the authors establish an interesting link between the image aesthetics and the visual sentiment, expressed in terms of valence and arousal.

Wang et al. \cite{Wang2013BeautyIH} introduce the inclusion of mid-level features in the form of classemes \cite{torresani2010efficient} apart from low-level image features and motion features for the aesthetic assessment of videos. Bhattacharya et al. \cite{Bhattacharya2013TowardsAC} propose analyzing aesthetics at different levels (cell, frame, and shot) while also being among the first to incorporate affective cues in the form of human sentiment classifiers from Citibank detectors \cite{Borth2013LargescaleVS}.  
The relationship between colors and the evoked emotion has also been studied \cite{Kim2016KeyCG}, linking features to perception. A closely related theme is understanding a venue's ambiance across 13 dimensions (e.g., loud, trendy, romantic) from the visual signal gathered from social media images \cite{Santani2014LoudAT}, something also explored using features from CNNs and Gradcams for localizing ambiance \cite{Can2018AmbianceIS}.

While such handcrafted features show promise, the emergence of CNNs for automatically learning features also motivated researchers to use them for this task, one of the early works being done by Lu et al. \cite{Lu2014RAPIDRP}. Schwarz et al. \cite{Schwarz2016WillPL} propose automatically learning the aesthetic space using feature vectors from CNN and relatively ranking them. The authors use the behavior signal in the form of user engagement as a weak proxy to train the model. This approach scores high on the AVA database. Using this, the authors demonstrate the capabilities of video to extract aesthetically pleasing frames from video sequences. Wu et al. \cite{wu2023exploring} propose to disentangle and separately learn the aesthetic (semantics and composition) and technical (blur, artifacts) aspects of video quality assessment.

Talon et al. \cite{talon2025seeing}, highlight the inability of Vision Language Models in understanding abstract concept words which are prevalent in fashion such as (\textit{airy, chic}). To tackle this authors try to map the shift from \textit{abstract→
concrete} concepts using PCA in the latent embedding space. This shows improvement from baseline zero-shot retrieval (31.1\% vs. 43.7\%). However this work is only focussed on images, we believe such approaches are vital to bridge gap between the abstract and concrete concepts. Wu et. al. \cite{Wu2023QAlignTL} propose Q-Align for learning discrete levels of aesthetics a deviation from previous score-based systems. The framework also unifies image and video quality assessment under one umbrella. Q-Bench-Video \cite{zhang2025q} is a comprehensive benchmark evaluating models on their aesthetic assessment capabilities.The benchmarks highlight a significant gap between human performance and the best performing model (58.7\% vs. 81.56\%). Models are shown to struggle with open-ended queries and detecting distortions in AI-generated content. 
Using Foundation models, we refer the reader to \cite{Zhang2024QualityAI} for a broad outline of works in a similar spirit. 

An interesting application of understanding human perception is understanding video content's memorability. This unifies different concepts like aesthetics, emotions, and generic video understanding. The pioneering work of quantifying image memorability from low-level features \cite{Isola2011WhatMA} lays the foundation to be applied to videos. Foundation models for visual understanding (EVA-CLIP \cite{Sun2023EVACLIPIT}), cognitive reasoning (verbalized scene descriptors), and world knowledge (LLM) can be leveraged for predicting video memorability, achieving state-of-the-art results on multiple memorability benchmarks as well \cite{HariniS2023LongTermAM}.

The progression of techniques in visual understanding has advanced from handcrafted features to deep convolutional neural networks and, more recently, to foundation and multimodal models. Aesthetics play a crucial role in shaping human perception, whether of an object, scene, or video. Foundation models, by connecting scene representations with abstract notions of beauty through natural language, enable a deeper understanding and more expressive descriptions of aesthetic qualities. This not only enhances the interpretability of aesthetics in video content but also paves the way for leveraging such models to increase the impact of visual media, given the established link between aesthetic appeal and video memorability. Furthermore, expert-driven aesthetic recommendations can be applied to optimize the visual appeal of spaces, presenting valuable opportunities in fields such as urban computing. 

\subsection{Intent}
Video intent is a complex and underexplored notion that may refer to, among other, the creator's or uploader's motivation for recording a video \cite{7123627}, user information need in video search \cite{10.1145/2393347.2396424} or even interactions between the semantic concepts in the video content. 

Here, we discuss the umbrella abstract concept of intent with different scenarios ranging from understanding action intent to understanding conversational and communicative intent. Since understanding intent requires us to understand not only the raw signal information but also a proper understanding of the context and sometimes even real-world grounding, we argue this task reaches higher levels of semantics than tasks like action recognition. The section highlights some important datasets and techniques from the image and video domains.

\textbf{\textit{Action Intent:}}
Epstein et al. \cite{Epstein2019OopsPU} present a dataset for predicting unintentional actions in a video that they crawl from YouTube fail compilations. Diagnostic annotations are also provided for a small subset of videos, telling the root cause of failure. Multiple sub-problems emerge from this dataset: classifying whether the action is intentional, localizing the transition from intentional to unintentional action, and finally anticipating whether the actions would lead to failure. The authors use self-supervised pretraining strategies on the dataset and find that training models to predict video speed effectively enhances performance. It is worth noting that the mentioned strategies perform well compared to transfer learning approaches (models trained on Kinetics and fine-tuned on the dataset). Li et. al.~ \cite{Li2023IntentQACV} propose a new VideoQA task of answering action intent and propose a Context-aware Video Intent Reasoning model (CaVIR) that understands action intent. The challenge comes from understanding the context and situation from a video. The dataset is constructed from the NExT-QA \cite{Xiao2021NExTQANP} dataset under the following concepts: causal why and causal how. Also, temporal sequences of actions are considered for achieving an objective. The approach utilizes situational context and GPT for common-sense context, showing how Foundation models can help bridge the existing common-sense gaps. 

Xie et al. \cite{Xie2023FunQATS} go a step further and propose a new benchmark (FunQA) for understanding surprising content, as they propose a new benchmark for training and testing models against their counter-intuitive comprehension abilities across humor, magic, and creativity: all inducing a positive surprise and challenging the normal action intent flow. The authors highlight that the benchmark is not trivial: humans achieve an 86\% error rate (most difficult being magic) by being shown randomly sampled frames, therefore underscoring the necessity for video-based reasoning. 
While only captions are enough for models like GPT-4V to score high on benchmarks like NExT-QA \cite{Xiao2021NExTQANP}, they score close to random when tested on this benchmark, highlighting the issue of memorisation for trivial benchmarks. 
Ablations show that while instruction-based models are better at reasoning than captioning models, all models collapse in the task of timestep localization in scenarios involving humor, magic, or creativity. Furthermore, models fail to capture temporal sequencing and incorrectly explain creative or magical elements in videos. Performance on the Multiple Choice variant of the dataset is shown in Table \ref{tab:performance_table}.

A similar direction is explored in \cite{chinchure2025black}, which augments the Oops! \cite{Epstein2019OopsPU} dataset and proposes three tasks that tackle forecasting, detecting, and reasoning for unintentional actions in fail/prank videos. Ablations show that forecasting is trivial for models, which the authors attribute to similar tasks in the training distributions of these Foundation models (0.78 for humans, the same as Gemini-1.5-Pro). However, models fail to capture fine-grained details, such as individuals and fast-paced events, leading to poor performance on other tasks. Chain-of-thought is shown not to help much, as wrong assumptions lead to incorrect predictions.

\textbf{\textit{Conversational Intent:} }Conversational intent arises when verbal communication combines with non-verbal cues and actions. This fusion makes it crucial to understand intent by analyzing all available modalities, including visual and audio cues. Xu et al. \cite{Xu2021Vid2IntDI} propose a new dataset and task, Vid2Int-Deception, for segment-level intent understanding in conversations. The authors use optical flow to capture subtle movements of the eyes and body alongside acoustic features, which show significant improvements compared to raw frames. An important dataset in this context comes from \cite{Zhang2022MIntRecAN} named MIntRec, curated from the TV show Superstore. The authors propose the problem of understanding intent from multimodal data such as video, audio, and subtitle text. The intent categories are curated broadly into two categories: \textit{Express emotions and attitudes} and \textit{Achieve goals}, which have further fine-grained annotations. Extensive ablations demonstrate that fusing modalities leads to an increase in the performance both in coarse and fine-grained intent classification. However, the performance is sub-optimal as compared to humans. The performance gap can be attributed to understanding essential visual (body movement) and audio cues (tone of voice), which are trivial for humans. The authors also expand the dataset to MInetRec 2.0 \cite{Zhang2024MIntRec20AL} with more videos (2,224 to 15,040), incorporating multi-turn conversations and considering out-of-scope utterances while also expanding on the intent categories (20$\rightarrow$30). The authors demonstrate the role of understanding multimodal signals by showing an increase in F1-score by 3.57 points by making the model multimodal as compared to only text ChatGPT's poor performance with text-only inputs compared to humans, who can perform better by a margin of 30\% when provided with 10 dialogue examples.

\textbf{\textit{Communicative intent}:} Communicative intent refers to a user's motivation to upload content to the Internet, aiming to convey emotions, ideas, or information to the audience. We cover this as a part of a broader concept of \textit{uploader intent} \cite{7123627} as discussed further. Kruk et al. \cite{Kruk2019IntegratingTA} explore meaning multiplication from social media posts, where multiple modalities combined create rich and abstract meanings. Combining text and visual features improves understanding of the intent behind social media posts. A similar study comes from  Jia et al. \cite{Jia2020IntentonomyAD}, who propose Intentonomy, a large-scale dataset with a broader taxonomy for understanding human intent from images to bridge the gap between computer vision and psychology. Experiments are conducted to understand the relationship between scenes and objects, which form context, with text and hashtags refining the interpretation. 
Yang et al. \cite{yang2025uncertain} propose the first benchmark MINE, which jointly models intent and emotion from social media posts spanning different modalities (text/image/video/audio). Visual modality is shown to improve performance relative to unimodal approaches. The proposed BEAR framework outperforms baselines; however, the scores remain modest, highlighting the room for improvement. No ablations are conducted using Foundation models for this work.

In summary, the study of intent covers a wide range, from analyzing the motivations behind specific actions to interpreting subtle cues in conversations and the complex messages found in social media. Grasping intent is essential for understanding human behavior. As discussed earlier, applying common-sense reasoning can significantly simplify interpreting actions and conversational intent. Foundation models offer valuable support for common-sense reasoning. However, interpreting socially nuanced behaviors and dialogues remains challenging and relatively unexplored since many actions are deeply rooted in particular cultures or communities. Additional context—such as external knowledge or cultural references—may be necessary to interpret such behaviors and their underlying intent. Examining the communicative intent behind social media posts is also a compelling area, as it provides insights into rhetorical strategies. Yet, relevant datasets are lacking, especially in the video domain. This highlights the ongoing need for research in multimodal reasoning, expanding datasets, and leveraging Foundation models to address these intricate semantic challenges.

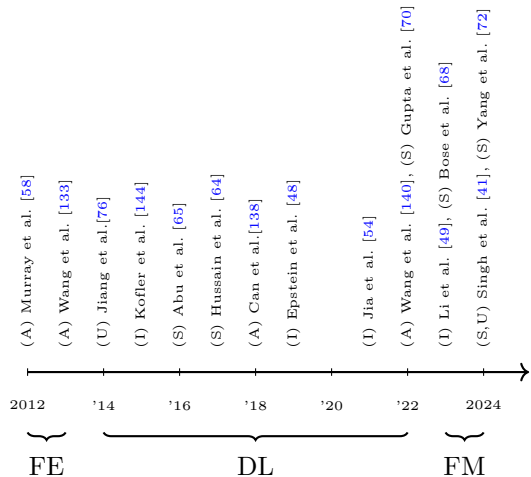
\begin{figure}[h]
    \centering
    \begin{tikzpicture}
    \def\StartYear{2012} 
    \def\EndYear{2025}   
    \def\bracketheight{1.5cm} 
    \def\FixedLength{6} 
    \def\mark{0.05}
    \pgfmathtruncatemacro{\YearDiff}{\EndYear - \StartYear}
    \pgfmathtruncatemacro{\Scale}{\FixedLength * 100 / \YearDiff} 
    \pgfmathsetmacro{\UnitScale}{\Scale / 100} 
    \draw[->, thick] (0,0) -- (\FixedLength+0.1*\FixedLength,0);
    \foreach \i in {0,2,...,\YearDiff} {
        \pgfmathsetmacro{\XPos}{\i * \UnitScale}
        \pgfmathtruncatemacro{\Year}{\StartYear + \i}
        \ifnum\i=0
            \draw (\XPos,-\mark) -- (\XPos,\mark) node[anchor=north, font=\tiny] at (\XPos,-.25) {\Year};
        \else
            \ifnum\i=\YearDiff
                \draw (\XPos,-\mark) -- (\XPos,\mark) node[anchor=north, font=\tiny] at (\XPos,-.25) {\Year};
            \else
                \pgfmathtruncatemacro{\ShortYear}{\Year - 100 * int(\Year / 100)}
                \edef\YearLabel{\ifnum\ShortYear<10 0\number\ShortYear\else\number\ShortYear\fi}
                \draw (\XPos,-\mark) -- (\XPos,\mark) node[anchor=north, font=\tiny] at (\XPos,-.25) {'\YearLabel};
            \fi
        \fi
    }
    \newcommand{\addpaper}[2]{ 
        \pgfmathtruncatemacro{\PubYear}{#1}
        \ifnum\PubYear<\StartYear
        \else
            \ifnum\PubYear>\EndYear
            \else
                \pgfmathsetmacro{\PubXPos}{(\PubYear - \StartYear) * \UnitScale}
                \edef\Angle{90}
                \node[rotate=\Angle, anchor=west, font=\tiny] at (\PubXPos,0.2) {
                #2
                };
            \fi
        \fi
    }
    
    \newcommand{\addera}[3]{ 
        \pgfmathtruncatemacro{\EraStartYear}{#1}
        \pgfmathtruncatemacro{\EraEndYear}{#2}
        \ifnum\EraEndYear<\StartYear
        \else
            \ifnum\EraStartYear>\EndYear
            \else
                \pgfmathtruncatemacro{\EffectiveStart}{\EraStartYear < \StartYear ? \StartYear : \EraStartYear}
                \pgfmathtruncatemacro{\EffectiveEnd}{\EraEndYear > \EndYear ? \EndYear : \EraEndYear}
                
                \pgfmathsetmacro{\EraStartXPos}{(\EffectiveStart - \StartYear) * \UnitScale}
                \pgfmathsetmacro{\EraEndXPos}{(\EffectiveEnd - \StartYear) * \UnitScale}
                \pgfmathsetmacro{\EraMidXPos}{(\EraStartXPos + \EraEndXPos) / 2}
                
                \draw[thick, decorate, decoration={brace, amplitude=0.1cm, mirror}] 
                    (\EraStartXPos,-0.8) -- (\EraEndXPos,-0.8);
                
                \node[anchor=north] at (\EraMidXPos,-1) {#3};
            \fi
        \fi
    }
    
        \addpaper{2012}{(A) AVA - Murray et al.}
        \addpaper{2013}{(A) Beauty is Here - Wang et al.}
        \addpaper{2014}{(U) Viral Video Style - Jiang et al. }
        \addpaper{2015}{(I) Uploader Intent- Kofler et al.}
        \addpaper{2016}{(S) Youtube-8M - Abu et al.}
        \addpaper{2017}{(S) Pitts Ads - Hussain et al.}
        \addpaper{2018}{(A) Ambiance in Venues - Can et al.}
        \addpaper{2019}{(I) Oops - Epstein et al.}
        \addpaper{2021}{(I) Intentonomy - Jia et al.}
        \addpaper{2022}{(A) Wang et al. , (S) 3Massiv -Gupta et al. }
        \addpaper{2023}{(I) IntentQA - Li et al., (S) MMAU - Bose et al. }
        \addpaper{2024}{(S,U) BLIFT - Singh et al. , (S) Yang et al.}
        \addpaper{2025}{(S) AdsQA - Long et al. , (I) MINE - Yang et al. }

        
    
    \addera{2012}{2013}{FE}
    \addera{2014}{2022}{DL}
    \addera{2023}{2025}{FM}
    
\end{tikzpicture}
    \caption{Timeline of seminal works and their approaches. Legend: (A) Aesthetics, (I) Intent, (S) Semantic Theme, (U) User Behavior Modeling, (FE) Feature Engineering, (DL) Deep Learning, (FM) Foundation Models. }
    \label{fig:perception_timeline}
\end{figure}

\subsection{Semantic Theme Understanding}

The semantic theme of a video refers to the main subject or topic characterizing a segment or the entirety of a video. Understanding video at the level of the semantic theme requires capturing the deeper, often implicit, message or mood conveyed through the combination of visual content, audio, speech, and metadata. As such, the task has been of utmost importance for facilitating access to large video archives. Namely, while for years the main focus of research community was on automatic video annotations formulated at low and intermediate semantic levels, ranging from colors, textures and shapes to semantic concepts, actions and events, such choice was partly motivated by a particularly high level at which semantic themes are formulated, which has long been unattainable by the available computer vision tools. In contrast, when assigning manual annotations to documentary videos for retrieval and re-use, the archivists often use topical labels, such as \textit{cultural identity}, \textit{immigration} and \textit{history} \cite{rudinac2012leveraging}. We believe that the advent of Foundation models, capable of reasoning at increasingly high semantic levels, created an opportunity for addressing this long-standing challenge.
In this survey, we focus on video classification tasks where the final label isn't immediately obvious, indicating the need for a longer temporal context to understand the content. The immediate application of semantic theme understanding is assigning tags to a large video corpus. Video annotation has been a thoroughly researched problem to enhance search relevance, such as in tagging task professional and tagging task WWW introduced by MediaEval benchmark \cite{10.1145/1991996.1992047}. Recent advancements have broadened this scope to include more complex and abstract tasks such as advertisement understanding and film genre classification.

Advertisements employ different persuasive strategies such that the meaning becomes more implied than obvious, making understanding semantic themes harder to grasp. We also explore open-ended social media datasets where video classification relies on interpreting the full sequence of actions rather than focusing on a single action to predict the video topic. Finally, we discuss other image-based datasets for semantic themes that tackle abstract concepts.

\textbf{\textit{Advertisements}:} In the seminal paper by Hussain et al. \cite{Hussain_2017_CVPR}, the authors propose a rich dataset consisting of both image and video advertisements, although the video advertisements (3477) are not plenty as compared to the image ads(64k). The dataset includes topic annotations, such as \textit{animal rights} and \textit{alcohol abuse} and sentiment (e.g., \textit{feminine} and \textit{amazed}) and question-answer pairs regarding topics. The authors also organize and propose a set of techniques that can be employed to understand advertisements automatically, from literal ads that can be decoded by simple object detection and text recognition to advertisements that use abstract concepts like symbolism, affective recipes like surprise and humor, atypical objects, and emphasizing the qualities of the product via physical processes. C3D network \cite{tran2015learning} is employed to extract features from videos, perform downstream classification tasks of topic and sentiment recognition, and predict whether the video is funny or exciting. The Tencent Ads Understanding dataset \cite{Wang2021OverviewOT} includes a wide range of annotations for a deep understanding of videos, including semantic themes, emotions, aesthetics, production styles, and storytelling structure, to name a few. The challenge combines temporal segmentation with multi-label classification over these attributes. Different transformer backbone architectures are employed to extract fused features for making predictions. 

A significant dataset in this field is presented by Bose et al. \cite{Bose2023MMAUTowardsMU}, which offers a series of challenging tasks. These tasks consist of (1) identifying the topic, utilizing an extensive topic taxonomy developed from various datasets; (2) detecting changes in tone, indicated as a binary variable that shows whether an advertisement shifts from a positive to a negative tone; and (3) determining if the video communicates a social message, which is also represented as a binary variable. The authors also propose a variety of baselines: one constituting video transcripts, which are then used for zero-shot reasoning tasks employing different LLMs, the other using exact features from text transcript using Whisper \cite{Radford2022RobustSR} and encoded via \cite{Devlin2019BERTPO}, audio: using Audio Spectogram Transformer \cite{Gong2021ASTAS}, and video: using keyframes extracted from PySceneDetect\footnote{\url{https://github.com/Breakthrough/PySceneDetect}} which are encoded using CLIP \cite{Radford2021LearningTV}. The authors conduct different interventions and demonstrate how performance varies by combining different modalities. For example, social message detection performs better when using text and vision modalities, while tone transition works better when using audio signals with vision. Fusing all three modalities generally works better than using uni-modal approaches. The difference between zero-shot LLM performance and the multi-modal approach becomes apparent in the topic categorization task, where the zero-shot performance of GPT4 falls behind by 32\% compared to the best-performing supervised multimodal model. This may be attributed to using only transcripts as input signals for LLMs compared to the multimodal inputs.

Recently, video understanding benchmarks have started to pivot on advertisements due to their complex narrative structure and involvement of multiple abstract concepts. AdsQA \cite{long2025adsqa} has QA annotations for theme and core message extraction, along with other aspects such as visual concepts, emotions, persuasion strategies, and potential audience modeling. The performance of all models overall is significantly behind the human performance (see Table \ref{tab:performance_table}). They do relatively well at identifying themes and target audiences from characters, scenes, and slogans, but struggle with deeper reasoning tasks like persuasion strategy mining (40.4\% vs. 29.3\% for Gemini-2.5-Pro \cite{comanici2025gemini}). Notably, longer chain-of-thought reasoning degrades performance, even lower than baseline, since ad reasoning doesn’t follow the typical \textit{if A, then B} logic seen in math or code tasks. Parallel research on VideoAds \cite{zhang2025videoads} reports similar issues, with poor performance on reasoning-heavy tasks compared to static visual recognition. The importance of multimodality is highlighted by an increase in performance for models, such as LLaVA-Video (from 66.58\% to 71.73\%). Vision is essential, as shown by GPT-4o’s weak results with text-only input. Discussing the edge cases, models struggle to maintain temporal coherence amid frequent scene changes. Interestingly, some open-source models outperform GPT-4o (see Table \ref{tab:performance_table}), while Gemini-2.5-Pro \cite{comanici2025gemini} remains the top performer.

\textbf{\textit{Social Media}:} Gupta et al. \cite{Gupta20223MASSIVMM} propose a dataset (3MASSIV) and also study micro-videos on social media platforms covering the concepts, such as \textit{philanthropy}, \textit{pranks} and \textit{romance} as labels, affective states, and the type of media content. Classifying videos into such concepts requires understanding objects, scenes, events, actions, and interactions. Baseline experiments demonstrate difficulty in understanding \textit{reaction videos}, a content form where users react and publish their reactions, as the models cannot focus on respective salient parts. Human-centric labels such as \textit{pranks} and \textit{fails} often get misclassified.

\textbf{\textit{Static Images and Abstract Concepts}:} Understanding semantic themes and abstract concepts in images has been a recurrent problem. Tasks like retrieval of images from abstract concept \cite{Cerini2024RepresentingAC} and generation of images representing abstract concepts \cite{Liao2023TexttoImageGF} are already gaining traction, showing the immense potential of Foundation models in the domain. Calabrese et al. \cite{Calabrese2020FatalityKT} propose a dataset linking images to the Wordnet tree \cite{Miller1995WordNetAL}, including non-concrete concepts. Efforts to understand deeper semantics of images using external knowledge either from Foundation models  \cite{Yang2024CanLM, Li2024MVPBenchCL} or knowledge graphs \cite{Kalanat2022SymbolicID, Aditya2019IntegratingKA} have proved to be effective. Artwork images require understanding cultural context \cite{githubGitHubTategallerycollection,Stefanini2019ArtpediaAN, Achlioptas_2021_CVPR, Garcia_2018_ECCV_Workshops,wang2025artragretrievalaugmentedgenerationstructured} where Foundation models present a unique opportunity to be applied.

We have identified tasks in both image and video domains that focus on interpreting deeper semantics to uncover the overarching theme of the content. This approach, however, extends beyond just advertisements and films, serving as a general framework applicable to various creative forms such as visual art, theatre, and more. Building systems capable of such semantic understanding requires Foundation models that can grasp cultural contexts, interpret subtle visual signals, and demonstrate common-sense reasoning. These capabilities should be
an addition to accurate object and scene recognition to ensure a robust pipeline. We also highlighted several benchmarks that are crucial for evaluating and enhancing such models. Looking ahead, research should aim to harness world knowledge embedded in Foundation models and integrate diverse aspects of abstract concept understanding across multiple modalities and domains.

\subsection{User Behaviour Modeling}
The viral potential of a video is an interesting problem requiring a deeper understanding of human perception at scale. While video virality has been a recurring research topic since the expansion of content sharing and social networking sites  \cite{shamma2011viral}, recent research has also linked the viral potential of a video to brain activity \cite{tong2020brain}. User interactions like comments, shares, and likes provide a weak proxy for a stronger perceptual signal like fMRI readings. Here, we explore how video content features can influence its likelihood of going viral.

Early attempts to understand the viral potential of videos on social media platforms use a multitude of approaches spanning from using metadata, popularity indicators, and simulating video propagation using graph networks \cite{Figueiredo2013OnTP, Pinto2013UsingEV, Li2013OnPP, Yu2014TwitterdrivenYV}. An essential piece of work in a similar direction comes from Jiang et al. \cite{Jiang2014ViralVS}, highlighting one of the first large-scale studies on the topic. Through statistical analysis, the most crucial feature determinants of a viral video are extracted and then used to train a modified HMM model to forecast the peak day of views for such videos. Although metadata was used extensively, the visual features from SIFT \cite{Lowe2004DistinctiveIF} were used only to filter duplicate videos from the dataset.

Deza et al. \cite{Deza2015UnderstandingIV} conducted one of the initial studies to understand the effect of visual content on virality. The authors highlight that low-level features would fail to predict relative virality, indicating the need to understand visual content on high-level semantics. Using deep learning models for extracting high-level features does show improvement \cite{AlamedaPineda2017ViraliencyPL}. However, models are still prone to misunderstanding cultural nuances as they lack context. Chen et al. \cite{Chen2016MicroTM} argue to learn a shared latent space representation for different modalities (social, visual, acoustic, and textual), thus enabling the prediction of video popularity. For the visual modality, the authors demonstrate the effectiveness of including high-level features such as aesthetics \cite{bhattacharya2013towards}, visual sentiment \cite{Borth2013LargescaleVS} over color histograms, and object features. Aligned in this direction, Fontanini et al. \cite{Fontanini2016WebVP} use the content features from CNN and visual sentiments from DeepSentiBank \cite{Chen2014DeepSentiBankVS} for a similar task. 

Recently, Foundation models have been evaluated on the large-scale Global Popular Video Dataset (GPVD) \cite{kayal-etal-2025-large}. The task involves classifying videos as either \textit{local hit} or \textit{global big hit} based on view counts and geographic spread. Claude Sonnet 3.5 \cite{anthropic2024claude35sonnet}, with hypothesis generation and supervised signals, achieves 85.5\% accuracy, outperforming supervised baselines at 80\%. The authors generate videos from text and demonstrate that these summaries perform much better than using only titles. However, there are still challenges, such as capturing cultural nuances and avoiding hallucinations in video-to-text conversion.

With the hypothesis generation capabilities, LLMs can be leveraged to predict the popularity of user-generated content, as shown in \cite{Zhou2024HypothesisGW}. Works like \cite{Bhattacharya2023AVI} showcase Foundation models' zero-shot video understanding capabilities on high-level semantic tasks. In contrast, \cite{Khandelwal2023LargeCA, Singh2024TeachingHB} demonstrate how Foundation models better understand content when fine-tuned on user engagement signals such as likes to views ratio and top 5 comments. A similar direction has been explored for images previously by Jia et al. \cite{Jia2021ExploringVE}, where user engagement signals are converted to pseudo-labels. This improves the model's performance on downstream tasks like understanding emotions, political leanings and hateful memes. Singh et al. \cite{Singh2024TeachingHB} fine-tune the LLaMA-Vid \cite{Li2023LLaMAVIDAI} model using the above signals on their curated dataset and demonstrate significant improvements on a variety of tasks dealing with high-level semantics, such as topic classification, sentiment understanding, persuasive strategy understanding and memorability prediction. This is on top of predicting the user behavior signal, i.e., the popularity. Expanding upon the capabilities of Foundation models, Lyu et al. \cite{Lyu2024GPT4VisionAA} thoroughly describe the scenarios where these models (GPT-4V) may be used as a social media analysis engine. This covers a range of problems, all involving a deeper understanding of content and going beyond factoid information while requiring \textit{world context}, all of which are done better using Foundation models.

\begin{figure}
    \centering
    \includegraphics[width=\linewidth]{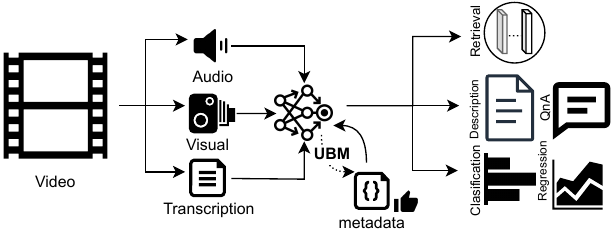}
    \caption{Directly forecasting user engagement indicators, such as the number of likes and top comments (UBM) during the training phase enhances effectiveness on subsequent tasks like classification, question answering, and recognition of abstract concepts as shown in \cite{Singh2024TeachingHB, Jia2021ExploringVE}.}
    \label{fig:enter-label}
\end{figure}

The abovementioned approaches suggest that this problem has been approached differently across distinct eras, evolving from basic feature extraction methods to leveraging multimodal Foundation models. Foundation models, which possess contextual, cultural, and semantic understanding, can be effectively applied to the challenge, as demonstrated in the concluding sections. Beyond predicting virality, a valuable direction would be understanding why a video gains popularity, where factors like memorability, emotional response, and visual appeal play key roles. Such insights could also help address the misuse of viral content for spreading misinformation or toxicity. This has practical implications for verifying the integrity of advertisements and political messaging, ensuring virality is not driven by harmful content. As Foundation models advance, especially when fine-tuned with social cues, they hold great potential for producing more accurate and generalizable predictions of content popularity within complex social media landscapes.

\section{Emotions and Social Signals}
Emotions are preconscious social expressions of feelings and affect influenced by culture \cite{Munezero2014AreTD}, inherently manifesting as Social Signals. The earliest works in this domain can be traced back to Charles Darwin's evolutionary perspective, which has further attracted various psychological paradigms, developing it further \cite{gendron2009reconstructing}. It is essential to note that the terms emotion, sentiment, feeling, and affect have rigorous definitions and mean different concepts \cite{Munezero2014AreTD}. However, these are discussed in Affective Analysis as an umbrella concept. The Circumplex model of affect \cite{posner2005circumplex, kuppens2013relation} states that all affective states arise from variation over two axes: valence (ranging from unpleasant to pleasant) and arousal (ranging from deactivated to activated). Hence, the feeling of ‘being excited’ is associated with high valence and arousal. On the contrary, the feeling of being ‘depressed’ is associated with both low valence and low arousal. Researchers often treat this as a regression problem, predicting continuous values for both valence and arousal. A similar treatment exists for understanding relationships \cite{traupman2009interpersonal} between two humans with the following axes: control (ranging from submissiveness to dominance) and affiliation (hostility to friendliness).
Social signals are communicative cues conveying meaning beyond literal content \cite{poggi2011social}. Social Signal Processing is a pretty recent study first proposed by Alex Pentland \cite{pentland2007social} focused on capturing non-verbal signals to understand the outcome of social interactions. Comprehending social situations requires many visual and acoustic signals, such as facial expressions, physiological responses, and group dynamics. Developing systems that automatically understand these signals and respond appropriately will enable us to create empathetic and adaptive human-centric video AI systems. 

Based on the literature survey pipeline, we have grouped literature focusing on emotions and social signals into the following sub-sections:

\begin{figure*}[t!]
    \centering
    \includegraphics[width=\linewidth]{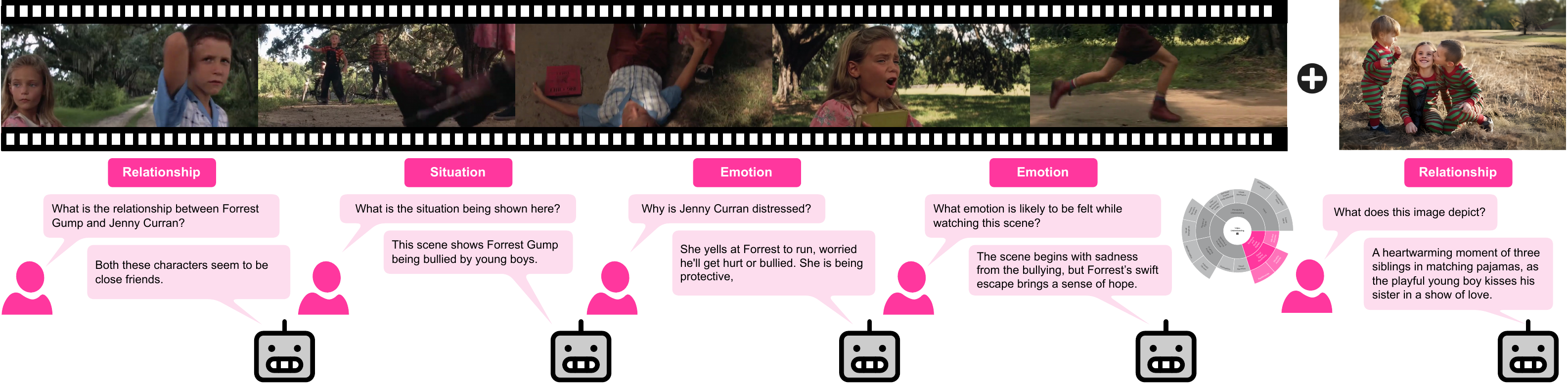}
    \caption{Pillar 2: Emotion and Social Signals emphasizes themes related to emotional expression and social cues, including understanding interpersonal relationships and situational context, as highlighted by an example from the MovieGraphs dataset \cite{Vicol2017MovieGraphsTU}. It also highlights the ability to infer relationships from static images, as demonstrated in SocialGPT \cite{Li2024SocialGPTPL}.}
    \label{fig:pillar2}
\end{figure*}

\begin{itemize}
    \item \textit{Affective Analysis}: Tasks related to understanding affect and emotions within the content and those induced in the recipient.
    \item \textit{Social Signal Processing}: Tasks focused on interpreting relationships and social situations, as commonly explored in video understanding. Our survey reviews these through the lens of social signal recognition.
\end{itemize}

Figure \ref{fig:pillar2} represents the idea that an ideal system should be able to understand the emotions of the actors and evoke emotions, and also process social situations from videos. Understanding this would require recognizing characters from different parts of a video, as well as understanding subtle features such as facial expressions and body language. Cinematic techniques, narrative structure, and background music can be used to intensify emotions, requiring multimodal understanding and temporal memory. We further discuss such cases in depth in the following sections.

\subsection{Affective Analysis}
Understanding emotions displayed and induced by the content is central to bridging the gap between raw signal information and human processing. Emotions form a central connection to many abstract concepts. They are significantly harder to map for visual signals as multiple visual concepts correspond to the same induced emotion and vice versa. Although the literature is vast, we look at some papers in this direction. For a detailed reference focusing on visual modality and its impact on perceived emotions, the readers are encouraged to look at \cite{Ortis2020ASO}. Further, we discuss how we capture emotions, ranging from classification to measuring valence and arousal, and how different objects, scenes, and actions are mapped to emotions.

One of the earliest works comes from Kang \cite{Kang2003AffectiveCD}, bridging the gap between low-level features like color, motion, and shots to high-level semantics like emotions. Hanjalic and Xu \cite{Hanjalic2005AffectiveVC} bridge the gap between psychophysiology and low-level audiovisual features.  The authors extract motion, shot changes, and acoustic as selected features to compute emotion-aware video analysis's arousal and valence curve, enabling emotion-based indexing and retrieval. 

Jiang et al. \cite{Jiang2014PredictingEI} were among the first to develop an automatic understanding of emotions in user-generated videos from YouTube and Flickr. Low-level visual features like HOG \cite{Dalal2005HistogramsOO} and SIFT \cite{Lowe2004DistinctiveIF}, acoustic features like spectral features, and attributes like Classemes \cite{torresani2010efficient} and  Sentibank \cite{Borth2013SentiBankLO} are employed to understand the evoked emotions from the video content. With developments in CNN and emotion datasets in the image domain, their applications to videos have also expanded. One of such directions aims to assign emotions of video frames to an auxiliary image dataset from \cite{Borth2013LargescaleVS}, CNN's being used as feature extractors and videos being treated as a Bag of Frames. Chen et al. \cite{Chen2016EmotionIC} propose to not only understand emotions from low-level features but also integrate events, scenes, and objects as a context. This improves performance compared to traditional deep features, closing the affective gap further. Among others, Xu et al. \cite{Xu2019VideoER} go further to leverage action concepts to understand emotions by filtering irrelevant concepts, demonstrating actions are also coupled with emotions, for e.g., \textit{celebrating} for joy and  \textit{smoking} for sadness. An interesting work also comes from Pang et al. \cite{Pang2020FurtherUV}, aiming to understand emotional adverbs (e.g., \textit{happily}, \textit{slowly}), which describe the associated mood behind an action. Facial features, motion features, context in the form of understanding scenes, and prior knowledge of related adverbs and actions are handy for this task.

Gestures form an essential part of human communication. Understanding gestures goes beyond action recognition and is vital in understanding the communicative intent. In \cite{Liu2021iMiGUEAI}, the authors propose iMiGUE for micro-gesture-based emotion analysis and propose an unsupervised S-VAE to learn latent representations from pose sequences, demonstrating the value of micro gestures in holistic emotion understanding. Li et al. \cite{Li2024EALDMLLMEA} demonstrate how Video-LLaMA can be leveraged to understand body language to convert emotions, surpassing supervised models. 

Mazeika et al. \cite{Mazeika2022HowWT} propose two large-scale datasets, mapping video to valence (via \textit{pleasantness} ranking) and emotions respectively, highlighting the need for AI systems to understand \textit{how content affects viewers}, instead of \textit{what is happening in the video}. Proposed tasks include predicting the emotion distribution and relative ranking according to pleasantness. Transformer-based action-recognition models trained on Kinetics \cite{Kay2017TheKH} can transfer to subjective tasks, showing that visual modality plays a vital role in understanding induced emotions. Models trained on this dataset perform better than CLIP with zero-shot performance; however, there is still scope for improvement. Ren et al. \cite{Ren2023VEATICVE} propose a new dataset for continuous valence and arousal prediction from videos incorporating both facial and contextual cues. Spatial and temporal context helps remove ambiguities in understanding the actors' emotions in the video, as confirmed by ablations conducted. Yang et al. \cite{Yang2023EmoSetAL} further push the research in this direction by proposing one of the largest image-based datasets (EmoSet) covering various visual concepts inducing different emotions.

Cheng et al. \cite{Cheng2024EmotionLLaMAME} fine-tuned Multimodal LLM (MLLM) and showed how audio, video, and text could be integrated for emotion recognition. This is important as the model needs to understand the subtle facial expression, temporal context, vocal tones, and auditory cues. The authors argue this is a significant challenge for models like GPT-4V. The authors propose a new dataset and use different encoders for respective modalities aligned with the LLaMA embedding space to solve this. Using this, the authors demonstrate superior performance compared to standard video MLLMs. More recently, AffectGPT \cite{lian2025affectgpt} proposes to pre-process via Q-Former the audio and visual modalities before sending to LLMs for processing. This reduces load on the LLMs and leads to better performance as shown in Table \ref{tab:performance_table}. Vanneste et al. \cite{Vanneste2024DetectingAE} propose an explainable AI pipeline that can not only detect emotions in the advertisements dataset \cite{Hussain_2017_CVPR} but also generate explanations as to which segment of the video contributes to the sentiment using LIME \cite{Ribeiro2016WhySI}. The approach, however, works on a single video frame, which is the key representative of the entire video. Explanations regarding the emotions are generated by perturbing the super-pixel regions of the keyframe, highlighting the essential sections contributing to the final prediction. Parallel research in video analysis involves identifying and ranking concepts that are understandable to humans, such as \textit{collision, object tracking}, etc., within the decision-making process \cite{kowal2024understanding}. Early layers concentrate on basic features, whereas the deeper layers represent more intricate events. Studies in this direction can be utilized to comprehend abstract concepts in videos via algorithmic methods or by developing layers that act as custom modifiers over low-level semantics to reach abstract levels of understanding.

Understanding induced emotions and emotions within the video content forms a central aspect of human perception in social contexts. The section highlighted different techniques to map actions, gestures, and facial expressions to emotions.  However, challenges remain in capturing subtle visual signals over long contexts. In order to create more reliable and broadly applicable models, future studies should concentrate on creating richer, multimodal datasets that capture the interaction between body language, scene context, audio cues, and facial emotions. Furthermore, incorporating explainable AI techniques will be critical for openness and confidence, especially in applications where comprehending the reasoning behind emotion forecasts is critical. Bridging the remaining affective gap between artificial systems and human emotional understanding will require resolving difficulties in cross-modal alignment, facilitating joint embeddings to capture subtle signals better, and temporal reasoning as big multimodal models continue to develop.

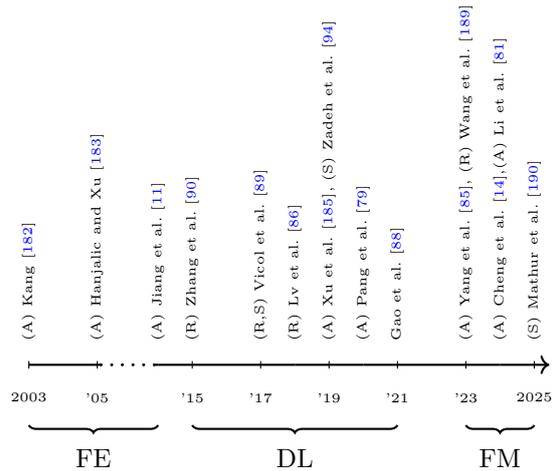
\begin{figure}
    \centering
    \begin{tikzpicture}

    \def\StartYear{2003}
    \def\EndYear{2025}
    \def\bracketheight{1.5cm}
    \def\FixedLength{10}
    \def\mark{0.05}
    \pgfmathtruncatemacro{\YearDiff}{\EndYear - \StartYear}
    \pgfmathtruncatemacro{\Scale}{\FixedLength * 100 / \YearDiff}
    \pgfmathsetmacro{\UnitScale}{\Scale / 100}

    \def\CompressedGap{0.8}  
    
    \pgfmathsetmacro{\posA}{0}                              
    \pgfmathsetmacro{\posB}{2 * \UnitScale}                 
    \pgfmathsetmacro{\posC}{\posB + \CompressedGap}         
    \pgfmathsetmacro{\posD}{\posC + 1 * \UnitScale}         
    \pgfmathsetmacro{\posE}{\posD + 2 * \UnitScale}         
    \pgfmathsetmacro{\posF}{\posE + 1 * \UnitScale}         
    \pgfmathsetmacro{\posG}{\posF + 1 * \UnitScale}         
    \pgfmathsetmacro{\posH}{\posG + 1 * \UnitScale}         
    \pgfmathsetmacro{\posI}{\posH + 1 * \UnitScale}         
    \pgfmathsetmacro{\posJ}{\posI + 2 * \UnitScale}         
    \pgfmathsetmacro{\posK}{\posJ + 1 * \UnitScale}         
    \pgfmathsetmacro{\posL}{\posK + 1 * \UnitScale}         

    \draw[thick] (\posA,0) -- (\posB,0);  
    
    \pgfmathsetmacro{\DotStart}{\posB + 0.05}
    \pgfmathsetmacro{\DotEnd}{\posC - 0.05}
    \pgfmathsetmacro{\DotSpacing}{(\DotEnd - \DotStart) / 5}
    \draw[thick] (\posB,0) -- (\DotStart,0);
    \foreach \i in {0,1,2,3,4,5} {
        \pgfmathsetmacro{\DotX}{\DotStart + \i * \DotSpacing}
        \fill (\DotX,0) circle (0.02);
    }
    \draw[thick] (\DotEnd,0) -- (\posC,0);
    
    \draw[thick] (\posC,0) -- (\posD,0);   
    \draw[thick] (\posD,0) -- (\posE,0);   
    \draw[thick] (\posE,0) -- (\posF,0);   
    \draw[thick] (\posF,0) -- (\posG,0);   
    \draw[thick] (\posG,0) -- (\posH,0);   
    \draw[thick] (\posH,0) -- (\posI,0);   
    \draw[thick] (\posI,0) -- (\posJ,0);   
    \draw[thick] (\posJ,0) -- (\posK,0);   
    \draw[thick] (\posK,0) -- (\posL,0);   
    \draw[->, thick] (\posL,0) -- ++(0.2,0); 

    \draw (\posA,-\mark) -- (\posA,\mark) node[anchor=north, font=\tiny] at (\posA,-.25) {2003};
    \draw (\posB,-\mark) -- (\posB,\mark) node[anchor=north, font=\tiny] at (\posB,-.25) {'05};
    \draw (\posD,-\mark) -- (\posD,\mark) node[anchor=north, font=\tiny] at (\posD,-.25) {'15};
    \draw (\posE,-\mark) -- (\posE,\mark) node[anchor=north, font=\tiny] at (\posE,-.25) {'17};
    \draw (\posG,-\mark) -- (\posG,\mark) node[anchor=north, font=\tiny] at (\posG,-.25) {'19};
    \draw (\posI,-\mark) -- (\posI,\mark) node[anchor=north, font=\tiny] at (\posI,-.25) {'21};
    \draw (\posJ,-\mark) -- (\posJ,\mark) node[anchor=north, font=\tiny] at (\posJ,-.25) {'23};
    \draw (\posL,-\mark) -- (\posL,\mark) node[anchor=north, font=\tiny] at (\posL,-.25) {2025};

    \node[rotate=90, anchor=west, font=\tiny] at (\posA,0.2) {(A) Kang};
    \node[rotate=90, anchor=west, font=\tiny] at (\posB,0.2) {(A) Hanjalic and Xu};
    \node[rotate=90, anchor=west, font=\tiny] at (\posC,0.2) {(A) Jiang et al. };
    \node[rotate=90, anchor=west, font=\tiny] at (\posD,0.2) {(R) Zhang et al. };
    \node[rotate=90, anchor=west, font=\tiny] at (\posE,0.2) {(R,S) MovieGraphs - Vicol et al. };
    \node[rotate=90, anchor=west, font=\tiny] at (\posF,0.2) {(R) Lv et al. };
    \node[rotate=90, anchor=west, font=\tiny] at (\posG,0.2) {(A) Xu et al., (S) SocialIQ - Zadeh et al.};
    \node[rotate=90, anchor=west, font=\tiny] at (\posH,0.2) {(A) Pang et al.};
    \node[rotate=90, anchor=west, font=\tiny] at (\posI,0.2) {(A) PERR - Gao et al. };
    \node[rotate=90, anchor=west, font=\tiny] at (\posJ,0.2) {(A) EmoSet - Yang et al. , (R) Wang et al.};
    \node[rotate=90, anchor=west, font=\tiny] at (\posK,0) {%
\begin{tabular}{l}
(A) EmotionLLaMA -- Cheng et al. \\
(A) EALD-MLLM -- Li et al.
\end{tabular}
};
    \node[rotate=90, anchor=west, font=\tiny] at (\posL,0.2) {(S) Social Genome - Mathur et al. , (S) Wang et al.};
    \pgfmathsetmacro{\EraMidOne}{(\posA + \posC) / 2}
    \draw[thick, decorate, decoration={brace, amplitude=0.1cm, mirror}] (\posA,-0.8) -- (\posC,-0.8);
    \node[anchor=north] at (\EraMidOne,-1) {FE};

    \pgfmathsetmacro{\EraMidTwo}{(\posD + \posI) / 2}
    \draw[thick, decorate, decoration={brace, amplitude=0.1cm, mirror}] (\posD,-0.8) -- (\posI,-0.8);
    \node[anchor=north] at (\EraMidTwo,-1) {DL};

    \pgfmathsetmacro{\EraMidThree}{(\posJ + \posL) / 2}
    \draw[thick, decorate, decoration={brace, amplitude=0.1cm, mirror}] (\posJ,-0.8) -- (\posL,-0.8);
    \node[anchor=north] at (\EraMidThree,-1) {FM};

\end{tikzpicture}

    \caption{Seminal works are marked along the timeline together with their approaches. Legend: (A) Affective, (R) Relationship, (S) Situation, (FE) Feature Engineering, (DL) Deep Learning, (FM) Foundation Models. }
    \label{fig:enter-label}
\end{figure}

\subsection{Social Signal Processing}
Social signal processing refers to the domain focusing on understanding and interpreting social signals in social contexts \cite{Vinciarelli2009SocialSP}. Social signals are intrinsically ambiguous and high-level semantic events \cite{Pantic2011SocialSP}, therefore offering a significant challenge as there is no direct correlation between visual and multimodal signals and the desired outputs. Signals like posture and interpersonal distance can reveal whether people are close friends, family, or strangers, and can indicate the intensity and nature of their relationship. Considering the core problems, \textit{who is interacting, how they relate, and in what context}, we dive into the issue of understanding relationships between characters and the social situations as a context. Further sections discuss both of them in detail.

\subsubsection{Relationships }
Inferring character relations is important to advance our understanding of social contexts and human behavior and analyze plots. This includes understanding cues like facial expressions, activities, and proximity between characters, requiring a deeper understanding of the content than usual action recognition. One of the early attempts to understand the relationship from facial signals comes from Zhang et al. \cite{Zhang2015LearningSR}, using the taxonomy provided by Kiesler \cite{Kiesler1983The1I}. The authors use a Siamese-like deep CNN to encode facial and spatial cues. Trained networks are also tested for ``Iron Man" to profile character relationships dynamically (\textit{competitive} vs. \textit{friendly}). Works like \cite{Li2017DualGlanceMF, Sun2017ADB} parallel propose understanding relationships in a social context; however, this is still limited to images.

Lv et al. \cite{Lv2018MultistreamFM} propose the first video dataset (SRIV) with social relation annotations (dominant, friendly, etc.). Spatial features from frames and motion features from flow are fused alongside audio features from the spectrogram to understand the relationships. Audio features help distinguish the type of relationships, e.g., a loud voice in arguments vs. a \textit{warm} relationship with a soft voice.  Liu et al. \cite{Liu2019SocialRR} progress the problem of understanding relationships in the video domain, highlighting how action, interactions, and storyline are essential for understanding relationships. They propose a new dataset (ViSR) addressing the limitation of previous datasets and also design a strong baseline capturing interactions between characters in a shot, tracking the same character over multiple shots, and recognizing objects in a setting to provide more context. However, a deeper understanding of the scene can help reduce ambiguity further. Gao et al. \cite{gao2021pairwise} propose recognizing emotional relationships from videos and introducing a large-scale multimodal dataset (PERR) to support this. The authors use multimodal fusion techniques to understand the type of relationship (such as \textit{hostile} or \textit{tense}) between characters. Baseline experiments involve integrating features such as facial expressions and body posture along with acoustic features such as background music and dialogues in the form of text via attention mechanisms and leveraging LSTMs \cite{hochreiter1997long} for temporal dependency.

A holistic understanding of social contexts and visual narratives often requires an understanding of relationships coupled with other indicators. One of the early attempts to understand relationships and situations in a social context (in the form of a Knowledge graph) comes from Vicol et al. \cite{Vicol2017MovieGraphsTU}  from videos and proposes a dataset (MovieGraphs) for the same. The dataset contains annotations for emotions, relationships, motivations, and interactions, covering various indicators for solving the problem. The authors demonstrate tight correlations with relationships, emotions, and situations, highlighting the need to understand the complete context and to reason better. Curtis et al. \cite{Curtis2020HLVUAN} propose HLVU, which aims to understand narrative through a knowledge graph, encapsulating relationships, interactions, etc. Relationships are explored via queries over the Knowledge Graph. A similar direction comes from Wu et al. \cite{Wu2021TowardsLV}, who propose an extended video understanding dataset (LVU) covering a wide range of tasks, including understanding relationships, interactions, speaking style, metadata based tasks such as recognizing the director and year, understanding the genre and also predicting the user engagement such as like ratio, etc. This is a challenging dataset and requires models to process information over a long sequence of frames efficiently. 

\begin{figure}[t]
    \centering
    \includegraphics[width=\linewidth]{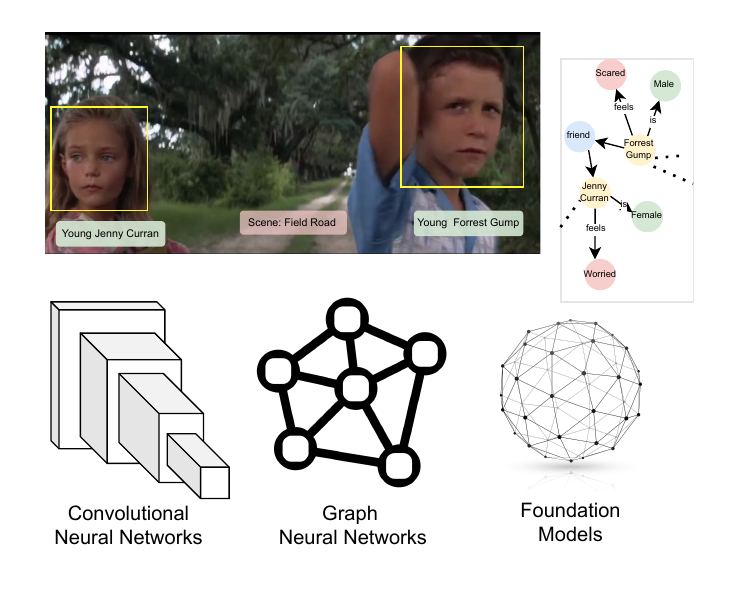}
    \caption{Understanding emotions, relationships, and social context benefits from visual cues like facial expressions, scene context, and structured annotations~\cite{Vicol2017MovieGraphsTU}. Training can therefore leverage CNN-based facial signal analysis~\cite{Zhang2015LearningSR}, graph-based methods~\cite{Wang2023ShiftedGA}, and external knowledge from Foundation Models~\cite{Li2024SocialGPTPL}.}
    \label{fig:enter-label}
\end{figure}

Wang et al. \cite{wang2025synergizing} tackle the problem of understanding relationships in long-term videos, highlighting how the relationship evolves over time. Facial and scene features constitute the visual signal, while BERT is used for dialogue embeddings, and long-term dependencies are captured by memory and update mechanisms. Characters are converted to graph nodes where the edges become the relationship type. Ablations across multiple datasets (MovieGraphs, ViSR, HLVU, and LVU) show the effectiveness of the approach. Further developments in Foundation models equip them with common-sense reasoning and, therefore, can be leveraged for understanding social contexts. Failure cases are attributed to co-variate shifts which could lead to LLMs not understanding the context in depth, for example \textit{opponent} in LLMs training distribution may include politics or sports as context while movies may have opponents based on moral differences. Works like SocialGPT \cite{Li2024SocialGPTPL} show the zero-shot capabilities of Foundation models to understand relations from images by leveraging the reasoning capabilities of LLMs. Dong et al. \cite{Dong2025KeyCG} demonstrate the capabilities of LLMs for enhancing understanding in the video domain by improving scores on the MovieGraphs dataset. 

Inferring relationships between characters is crucial for understanding social behaviors and can significantly benefit the automatic understanding of narratives and storytelling. The challenge remains in learning a detailed visual representation in a latent embedding space that captures subtle cues to resolve ambiguities, like facial expressions and postures, and audio cues that capture the tone. Scene representations also provide context, which helps determine the kind of relationship. For example, an office may exemplify a \textit{professional} relationship more than a \textit{romantic} one. Long-term dependencies in videos do offer a significant challenge in encoding these signals. Developments in character-centric video annotations and Foundation models to learn long-term dependencies would help advance this study further.

\subsubsection{Situation Analysis }
We look at the progress of systems toward social intelligence, going beyond just labeled sentiments. Socially intelligent systems should aim to navigate ambiguity, understand nuanced signals, comprehend multiple perspectives, and also adapt to feedback \cite{Mathur2024AdvancingSI}. This is still a relatively new challenge and calls for developing more datasets, which we actively advocate for. In the meantime, we explore existing research in multimodal video settings and some purely language-based approaches. We hope that more complex datasets will emerge to drive further progress in this area.

Zadeh et al. \cite{Zadeh2019SocialIQAQ} pioneer this research by proposing a multimodal QA dataset (Social-IQ) on videos from various social situations. The paper examines scenarios that involve analyzing conversations and actions. These social situations are complex, necessitating an understanding of not only the dialogue but also the interpretation of visual and emotional cues. It highlights how emotions evolve within a group in a specific context. Authors highlight that performing well on this benchmark requires a deeper understanding of context and common sense, something to keep in mind while evaluating advancements in Foundation models. Guo et al. \cite{Guo2023DeSIQTA} highlight the issues of bias in the dataset Social-IQ and showcase how models can employ shortcuts to achieve high accuracy without understanding context or input questions. This is demonstrated by incorrect answers clustering in the same region of space, a similar case for correct answers, forcing the model to choose correct answers. This highlights issues in constructing datasets of such type where biases creep in. Thus, the authors replace the incorrect option from each question with the correct answer of a different question, making the dataset more challenging. Injecting external modules for common-sense reasoning like VisualCOMET \cite{Park2020VisualCOMETRA}
proved to improve the score on this benchmark \cite{Natu2023ExternalCK}. Recent video benchmarks like Social Genome \cite{Mathur2025SocialGG} show that Foundation models are behind human performance, struggling with understanding subtle behaviors (e.g., lip movements) and implicit cues, indicating the shortcuts these Foundation models employ to answer. Li et al. \cite{Li2025VEGASTV} take a step in this direction and show how frame-sampling and emotion-aware fine-tuning strategies can benefit Foundation models by avoiding shortcuts.

Though Foundation models prove to be vital in understanding context and providing common-sense grounding, works like \cite{Mou2024AgentSenseBS, Zhou2023SOTOPIAIE} via extensive ablations over textual modality, demonstrate that these models still struggle with conflict resolution and understanding complex group dynamics. Foundation models underperformed in private information reasoning and violated social norms, highlighting the growth areas for more socially robust and grounded Foundation models. A relatively underexplored area is video memes, in contrast to image memes having a pool of datasets and literature \cite{Nguyen2024ComputationalMU}. Memes are often challenging as understanding their viral potential and evoked emotions, including detecting humor and toxicity, would require models to have a cultural and societal grounding. This can emerge as a new benchmark against which to test Foundation models.

Although this field is still emerging and needs more research, benchmarks beyond simple classification already offer significant challenges for testing Foundation models in these contexts. Social intelligence involves more than just recognizing emotions—it also demands common sense reasoning and an intuitive grasp of social dynamics, or the ability to \textit{read the room}, which is deeply human. Developing AI systems with such social awareness has broad applications and can help prevent misunderstandings in human-AI interactions. Benchmarks should encompass more subtle yet impactful situations, such as understanding racism and biases, as well as respecting cultural nuances. Research indicates that large language models (LLMs) encounter challenges in conflict resolution and maintaining privacy \cite{Mou2024AgentSenseBS}. Enhancing these foundation language models is a crucial first step. Another significant challenge is the creation of a unified space for language and vision, which would enable the capture of nuanced visual and linguistic cues in a shared latent embedding space. This remains an open and important area of research, with the potential to greatly advance the development of socially intelligent AI systems.

\section{Narrative and Rhetoric Analysis}
\begin{figure*}[ht!]
    \centering
    \includegraphics[width=\linewidth]{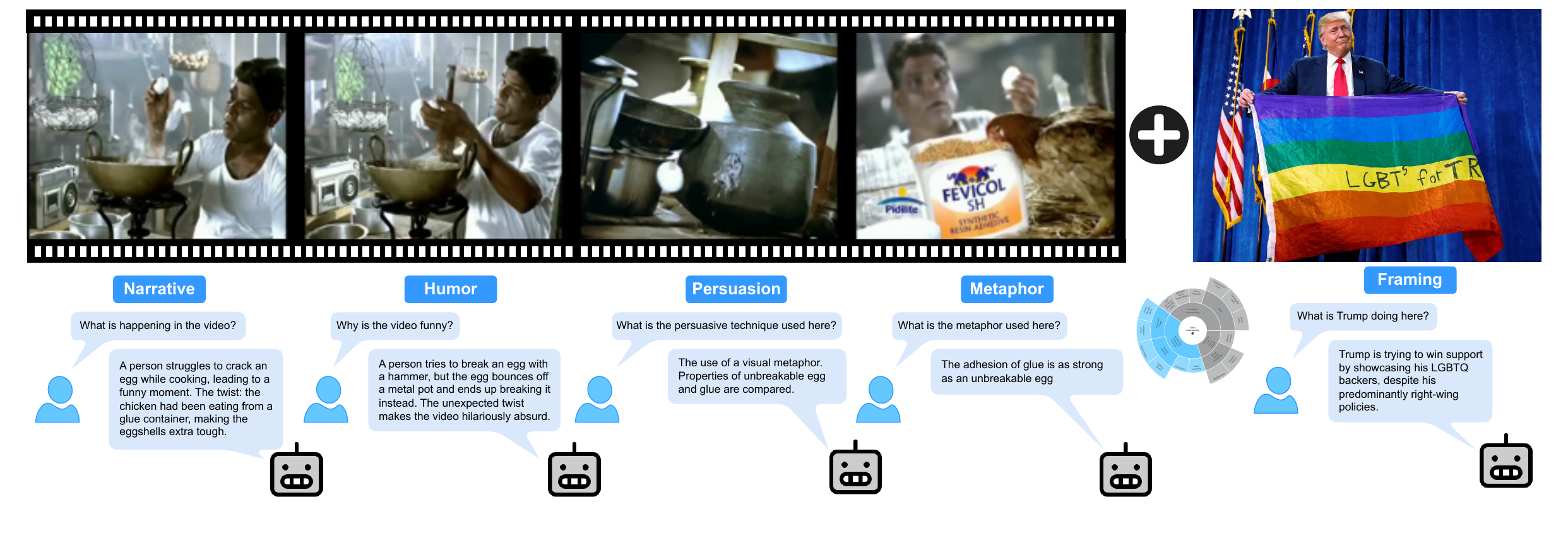}
    \caption{Pillar 3: Narrative and Rhetorical Analysis highlights key themes, such as interpreting visual narratives, recognizing figures of speech, analyzing persuasive techniques, and understanding political framing. Illustrative examples of Visual Metaphors are drawn from the VMC dataset \cite{kalarani2024unveiling}, and framing from \cite{Thomas2019PredictingTP}.
}
    \label{fig:pillar3}
\end{figure*}

Narrative and rhetorical analysis form complementary frameworks for understanding messages in communication literature. Narrative Analysis focuses on interpreting human experience through storytelling \cite{Ntinda2019NarrativeR} while Rhetoric Analysis evaluates elements to determine the persuasive impact and the strategic choice of language  \cite{marcellino2015revisioning}. Studies related to these can be traced back to Aristotle's work on persuasion and poetics \cite{kennedy2009new, bal2009narratology}.

Video content in movies forms a vital part of the visual narrative. Meaning is often communicated beyond the literal level through various figures of speech, such as metaphors, irony, and sarcasm. Various persuasive strategies are employed in video content to deliver messages different from their literal interpretation, such as in advertisements and political campaigns. Understanding such forms of content is vital to generalizing video understanding capabilities, enabling more profound insight into public discourse and creative media and such as films. Further sections discuss the video and image understanding research dealing with these distinct yet intertwined fields:

\begin{itemize}
    \item \textit{Visual Narrative Understanding}: Tasks related to understanding visual narrative from visual content such as films and comics.
    \item \textit{Figures of Speech}: Tasks related to understanding indirect forms of communication, going beyond literal depth. We include discussions on metaphors, humor, and sarcasm.
    \item \textit{Persuasion}: Tasks related to understanding various persuasive strategies used by advertisements and political campaigns.
    \item \textit{Framing Analysis}: Tasks involving the interpretation of opinions, political biases, and misinformation fall under Framing Analysis, a domain closely linked to Rhetoric Analysis \cite{kuypers2010framing}. 
\end{itemize}

Figure \ref{fig:pillar3} shows a clever ad that claims an egg is \textit{unbreakable} because the chicken ate an adhesive. This is a visual metaphor, where product traits are shown through analogy. To understand it, a model needs to follow the sequence of events, pick up on the humor, recognize the correct set of objects (e.g., ignoring other objects in focus, such as a pan or container), and understand how the entities relate to each other. This makes the task much more complex than standard video understanding. We also provide examples with political leanings, which require models to understand the long-term political, social, and cultural context to identify the issue being presented. An ideal model should be able to reason at these semantics levels while also maintaining the capability to reason over longer narrative structures, understand plot twists, climaxes, endings, and resolutions. In the subsequent sections, we discuss this in more depth.

\subsection{Visual Narrative Understanding}
Computational understanding of storyline and narrative poses a significant challenge as context is dispersed widely across time. A robust understanding requires systems to interpret context at the frame level while linking information across various hierarchical levels—from individual frames to shots, scenes, and the overall narrative. We examine understanding narratives and storylines from multiple perspectives, from standard comprehension and answering to understanding cinematic styles.

Addressing the semantic gap between the low-level features and the high-level human cognition, one of the first works to address computational understanding of movie content comes from Liu et al. \cite{Liu2008AHF}. The authors aim to answer questions like \textit{Who, What, Where, How}, going beyond raw features. Low-level features (like motion intensity, audio energy, sentence length, word types, etc.) are translated into information metrics using the Weber-Fechner Law \cite{hecht1924visual} for understanding human perception. They are further processed for information extraction, answering the questions above. Rohrbach et al. \cite{Rohrbach2015ADF} introduce the first large-scale movie dataset (MPII Movie Description) aligned with scripts and audio descriptions as captions. The approach combines feature extraction (Dense Trajectories for actions and CNNs as object and scene detectors) to predict semantic tuples (subject, verb, object, location), which are then used to caption movies. With this, the authors propose the possibility of understanding storylines and plots at a large scale. Along similar lines, Zhu et al. \cite{Zhu2015AligningBA} introduce a dataset and a task to align movies with their rich story explanations and book descriptions. The authors propose to retrieve the relevant movie description from the book's corpus given a movie scene/visual description. Multiple feature extractors like GoogleNet for frame-level features, LSTM and GRU for encoding descriptions, Context-aware CNN for combining similarity measures between visual and audio modality, etc., are used to ground visual information to the highly descriptive text in books. The dataset descriptions are still factoid-based but address the need to understand plots and context over time. Tapaswi et al. bring the understanding of plots to a question-answering domain and propose the dataset (MovieQA) for the same \cite{Tapaswi2015MovieQAUS}. Due to the field being in a nascent stage, text modality still performed better on the benchmarks than video-based features. However, as baseline experiments showed, temporal reasoning remained challenging, motivating research to understand and retain context over the temporal axis while integrating complex reasoning. Addressing the limitations of previous datasets and focusing more on characters rather than facts and events, Choi et al. \cite{Choi2020DramaQACV} propose DramaQA with character-centric annotations, including emotions and behaviors. The questions have various hierarchical levels inspired by human cognition, ranging from simple cue-based answering to high-level reasoning for causality. Multiple feature extractors, such as ResNet-18 for vision, GLoVe for emotion, and BiLSTM for temporal context, are used for different modalities, and they are then fused to answer questions. Context modules are required to ground the response to the evidence in the video by attending to relevant segments, motivating a deeper narrative understanding of this problem. Works like \cite{Chung2023LongSS} demonstrate the zero-shot question-answering capabilities of LLMs along with the captioning capabilities of models like BLIP to handle long-form question-answering, demonstrating how Foundation models can be leveraged to understand long videos. The challenge remains in the modality gaps and the ability to capture fine-grained details by image captioning models to convert subtle features into text.

Ye et al. \cite{ye2018story} enhance the video ads dataset \cite{Hussain_2017_CVPR} with climax annotations and introduce climax detection to aid sentiment recognition in video ads. The importance of narrative progression is emphasized by the authors. For instance, two ads might open with kids, conveying a sense of youth and optimism. But their plots are different. The first ad uses happy environments, like a toy store, to keep things positive. The second ad, on the other hand, shifts to unfavorable settings, such as a hospital room or basement, and ends with a sense of alarm in the closing frames. Using Freytag’s Pyramid, they segment videos into setup, action, climax, and resolution, identifying climaxes through peaks in audio, optical flow, and shot changes. Sentiment is derived from frame-level features like scenes, objects, and facial expressions. Their model outperforms baselines through feature permutations and ablation studies, showing that deep video understanding requires more than action analysis—it depends on a range of contextual cues. The first works on a holistic understanding of films come from Huang et al. \cite{Huang2020MovieNetAH}, who propose the first comprehensive movie understanding dataset (MovieNet) containing different annotations like character identities, cinematic styles, scene boundaries, and aligned descriptions, encompassing movies, trailers, photos, scripts, etc. Some of the highlighted tasks of this benchmark are genre classification, cinematic style prediction, shot scales and camera movements, and story retrieval, aligning synopses to relevant movie parts. Multiple deep learning models like ResNet and I3D extract the visual features and perform the tasks. This is an interesting benchmark for testing different models on a comprehensive understanding of cinematic content and can act as a bridge to advance video understanding.

Ghermi et al. \cite{Ghermi2024LongSS} benchmark Foundation models for long video understanding, highlighting several issues. Since Foundation models are trained with large-scale data, there is a high chance of data leaks. Therefore, there is a need to understand videos at a story level and create new datasets where the models are not trained. The authors make a benchmark with a question-answering framework and MCQs. The dataset ensures that the answers cannot be given just by reading the film's title, forcing the models to understand the plot in detail. The LLMs score poorly on the dataset compared to commercial films, indicating a data leakage, and language modality remains the dominating factor through subtitles, motivating research for visual modalities.

An effective way of visual storytelling to evoke emotions in manipulating shots \cite{Mercado2013TheFE, Achanta2006ModelingIF} and benchmarks exist to understand the cuts used in films \cite{Argaw2022TheAO, Pardo2021MovieCutsAN}.
An interesting application is to understand the visual objectification of characters or the \textit{male gaze} in movies, perpetuating gender biases \cite{Tores2024VisualOI}. Certain shots convey concepts, such as low-angle shots showing power dynamics or zooming on a body part showing dehumanization. Understanding long-term context and frame-level high abstractions is essential to understanding the task completely. Other features are also integrated alongside for more context, such as clothing, emotions, activity, etc. The authors use X-CLIP \cite{Ni2022ExpandingLP} to extract the video features and treat the problem as a binary classification.

Choi et al. \cite{Choi2016VideoStoryCV} propose the problem of sequencing clips to form a uniform storyline. The storyline consists of exposition, rising action, climax, and resolution. Optical flow is used to measure the video's activity and the plot stage (e.g., less optical flow corresponding to calmer scenes vs. climax with the high optical flow ). The sequence of videos is then optimized using a Branch and Bound algorithm. A similar interesting solution to the problem of sequencing shots during concerts using the audio modality by Wei et al. \cite{Wei2018SeethevoiceLF}. This is done by combining audio features at different semantic levels with a film language model, thus capturing cinematic techniques.

An interesting case opens to understanding and matching the background music with the ongoing visual narrative, matching the mood, genre, and theme. This is an interesting problem as it requires understanding high levels of semantic information from two modalities (vision and audio). Surs et al. \cite{Surs2022ItsTF} address the issue of matching the music with the artistic intent of the video, treating this as a cross-modal retrieval problem. CLIP and DeepSim are used to align in a common space. An interesting case study by the authors reveals the cultural biases that the model learns across various visual attributes (e.g., Black people are associated with `hip-hop,' etc.). Dong et al. propose MuseChat \cite{Dong2023MuseChatAC}, a framework to recommend interactive music based on visuals. This framework integrates LLMs for explainability and straightforward interpretation. Video frames and text are encoded with CLIP, and the audio is encoded via an AST encoder. All modalities are aligned in a common space, and the chat interface is the Vicuna model. This demonstrates the capabilities of Foundation models in bridging the gap between two modalities via natural language with abstract descriptions. The reasoning capabilities also help understand the pipeline's choices in the process, highlighting the model's agentic capabilities.

A closely related work is to understand the visual narrative from image sequences. Such forms are common in visual media such as comics, and understanding requires understanding the changes between consecutive frames and a robust knowledge of implicit actions taken. One such work in this direction comes from Wang et al. \cite{Wang2024MementosAC}, who benchmark MLLMs over this challenging task. The ablations reveal massive hallucinations in objects and behavior hallucinations, which become a challenge, and the narrative changes because of this. This reveals challenges in understanding changes.

Understanding storylines computationally continues to be a complex challenge, focusing on connecting basic visual elements with advanced human cognition. The field has evolved towards more complex tasks such as character reasoning, climax detection, and analyzing cinematic style, marking a transition from simple fact-based understanding to a more comprehensive grasp of narratives. Foundation models exhibit potential for analyzing long-form videos but encounter challenges like differences in modalities and issues with data leakage. As the field progresses, cleaner benchmarks must be developed, emphasizing less reliance on textual modality in the form of subtitles. A significant challenge is understanding long-form video narratives, which require storing and processing context over long temporal ranges. Progress in this area depends on creating efficient Foundation models optimized for processing long-range sequences.

\begin{figure}[h]
    \centering
    \begin{tikzpicture}

    \def\StartYear{2008}
    \def\EndYear{2025}
    \def\bracketheight{1.5cm}
    \def\FixedLength{9}
    \def\mark{0.05}
    \pgfmathtruncatemacro{\YearDiff}{\EndYear - \StartYear}
    \pgfmathtruncatemacro{\Scale}{\FixedLength * 100 / \YearDiff}
    \pgfmathsetmacro{\UnitScale}{\Scale / 100}

    \def\CompressedGap{0.8}  
    
    \pgfmathsetmacro{\posA}{0}                              
    \pgfmathsetmacro{\posB}{\posA + \CompressedGap}         
    \pgfmathsetmacro{\posC}{\posB + 1 * \UnitScale}         
    \pgfmathsetmacro{\posD}{\posC + 1 * \UnitScale}         
    \pgfmathsetmacro{\posE}{\posD + 1 * \UnitScale}         
    \pgfmathsetmacro{\posF}{\posE + 2 * \UnitScale}         
    \pgfmathsetmacro{\posG}{\posF + 1 * \UnitScale}         
    \pgfmathsetmacro{\posH}{\posG + 1 * \UnitScale}         
    \pgfmathsetmacro{\posI}{\posH + 1 * \UnitScale}         
    \pgfmathsetmacro{\posJ}{\posI + 1 * \UnitScale}         
    \pgfmathsetmacro{\posK}{\posJ + 1 * \UnitScale}         

    \pgfmathsetmacro{\DotStart}{\posA + 0.05}
    \pgfmathsetmacro{\DotEnd}{\posB - 0.05}
    \pgfmathsetmacro{\DotSpacing}{(\DotEnd - \DotStart) / 5}
    \draw[thick] (\posA,0) -- (\DotStart,0);
    \foreach \i in {0,1,2,3,4,5} {
        \pgfmathsetmacro{\DotX}{\DotStart + \i * \DotSpacing}
        \fill (\DotX,0) circle (0.02);
    }
    \draw[thick] (\DotEnd,0) -- (\posB,0);
    
    \draw[thick] (\posB,0) -- (\posC,0);   
    \draw[thick] (\posC,0) -- (\posD,0);   
    \draw[thick] (\posD,0) -- (\posE,0);   
    \draw[thick] (\posE,0) -- (\posF,0);   
    \draw[thick] (\posF,0) -- (\posG,0);   
    \draw[thick] (\posG,0) -- (\posH,0);   
    \draw[thick] (\posH,0) -- (\posI,0);   
    \draw[thick] (\posI,0) -- (\posJ,0);   
    \draw[thick] (\posJ,0) -- (\posK,0);   
    \draw[->, thick] (\posK,0) -- ++(0.5,0); 

    \draw (\posA,-\mark) -- (\posA,\mark) node[anchor=north, font=\tiny] at (\posA,-.25) {2008};
    \draw (\posB,-\mark) -- (\posB,\mark) node[anchor=north, font=\tiny] at (\posB,-.25) {'14};
    \draw (\posD,-\mark) -- (\posD,\mark) node[anchor=north, font=\tiny] at (\posD,-.25) {'16};
    \draw (\posG,-\mark) -- (\posG,\mark) node[anchor=north, font=\tiny] at (\posG,-.25) {'20};
    \draw (\posI,-\mark) -- (\posI,\mark) node[anchor=north, font=\tiny] at (\posI,-.25) {'22};
    \draw (\posK,-\mark) -- (\posK,\mark) node[anchor=north, font=\tiny] at (\posK,-.25) {2024};

    \node[rotate=90, anchor=west, font=\tiny] at (\posA,0.2) {(N) Liu et al.};
    \node[rotate=90, anchor=west, font=\tiny] at (\posB,0.2) {(Fr) Park et al.};
    \node[rotate=90, anchor=west, font=\tiny] at (\posC,0.2) {(N) MovieQA - Tapaswi et al. , (FoS) Chandrasekaran et al. };
    \node[rotate=90, anchor=west, font=\tiny] at (\posD,0.2) {(P) Siddiquie et al. , (Fr) MOSI - Zadeh et al. };
    \node[rotate=90, anchor=west, font=\tiny] at (\posE,0.2) {(P) ADVISE - Ye et al. };
    \node[rotate=90, anchor=west, font=\tiny] at (\posF,0.2) {(Fr) Politics of Image - Thomas et al.};
    \node[rotate=90, anchor=west, font=\tiny] at (\posG,0.2) {(N) DramaQA - Choi et al. , (N) HLVU - Curtis et al. };
    \node[rotate=90, anchor=west, font=\tiny] at (\posH,0.2) {(FoS) MultiMET - Zhang et al. , (Fr) SnchezVillegas et al. };
    \node[rotate=90, anchor=west, font=\tiny] at (\posI,0.2) {(N) Surs et al. , (FoS) Kumar et al.};
    \node[rotate=90, anchor=west, font=\tiny] at (\posJ,0.2) {(Fr) Sung et al.};
    \node[rotate=90, anchor=west, font=\tiny] at (\posK,0.2) {(N) Tores et al. , (FoS) Kalarani et al.};
    \node[rotate=90, anchor=west, font=\tiny] at (\posK+0.3,0.2) {(P) PALADIN - Liu et al. };

    \node[rotate=90, anchor=west, font=\tiny] at (\posK,0.2) {};

    \pgfmathsetmacro{\EraMidOne}{(\posA + \posB) / 2}
    \draw[thick, decorate, decoration={brace, amplitude=0.1cm, mirror}] (\posA,-0.8) -- (\posB,-0.8);
    \node[anchor=north] at (\EraMidOne,-1) {FE};

    \pgfmathsetmacro{\EraMidTwo}{(\posC + \posI) / 2}
    \draw[thick, decorate, decoration={brace, amplitude=0.1cm, mirror}] (\posC,-0.8) -- (\posI,-0.8);
    \node[anchor=north] at (\EraMidTwo,-1) {DL};

    \pgfmathsetmacro{\EraMidThree}{(\posJ + \posK) / 2}
    \draw[thick, decorate, decoration={brace, amplitude=0.1cm, mirror}] (\posJ,-0.8) -- (\posK,-0.8);
    \node[anchor=north] at (\EraMidThree,-1) {FM};

\end{tikzpicture}
    \caption{Seminal works are marked along the timeline with their approaches. Legend: (N) Narrative, (P) Persuasion, (Fr) Framing, (FoS) Figures of Speech, (FE) Feature Engineering, (DL) Deep Learning, (FM) Foundation Models. }
    \label{fig:perception_timeline}
\end{figure}
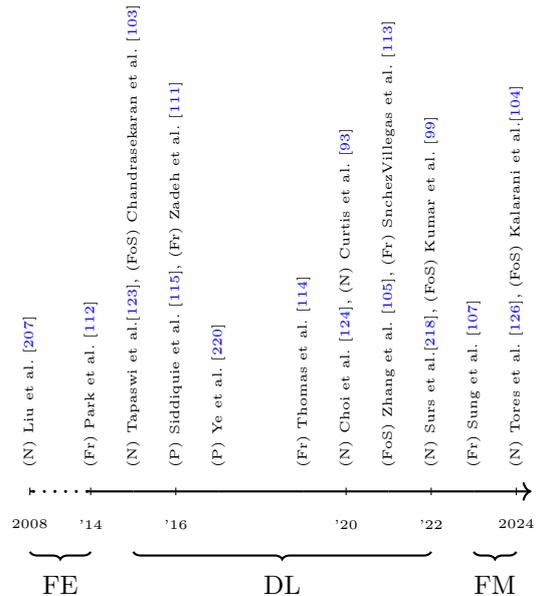

\subsection{Figures of Speech}
\textit{\textbf{Visual Metaphors}}: Metaphors form a large part of figurative language and are frequently used as a subtle means of communication in visual media such as films and commercials \cite{Forceville2024IdentifyingAI}. Defining metaphors as `understanding and experiencing one kind of thing in terms of another' \cite{norvig1985metaphors} is a good starting point for understanding and interpreting metaphors. Visual metaphors can occur at the frame level or with action sequences over multiple frames, highlighting the need for advanced video understanding. While understanding metaphors is not a new problem in the language domain \cite{Gao2018NeuralMD}, visual modality poses a different challenge. Even fewer are video datasets in the domain.

The first attempts at computational understanding of metaphors with visual modality in videos come from Alnajjar et al. \cite{Alnajjar2022RingTB}. The authors highlight the lack of datasets in this domain and propose the first multimodal dataset to understand metaphors. They also discuss how certain visual and acoustic elements communicate a message, such as flying money, indicating becoming rich. However, text modality still performs the best according to the experiments conducted, but this can be attributed to the small dataset size.

An essential extension to the ads dataset proposed by Hussain et al. \cite{Hussain_2017_CVPR} also comes from \cite{kalarani2024unveiling}, where the video advertisements are annotated with metaphors used and release a new task called the Video Metaphor Captioning (hereby referred to as VMC). Metaphors in advertisement use a mechanism where the \textit{primary concept} is linked to the \textit{secondary concept} via a common property. The authors operate in a low data regime and use GIT \cite{Wang2022GITAG} and Vicuna 13B \cite{Zheng2023JudgingLW} as their choice of models. Since the task is multimodal and bridging the gap between the vision and text modalities using the models requires training on a large corpus, the authors divide the training into two stages: pre-training on a synthetic video dataset and fine-tuning on VMC. While the authors were able to match the performance of SOTA models with this, all the models used in the study `lack a deeper understanding' of VMC. 

Advertisement campaigns use different objects in symbolism to convey a message. A gun may signal danger, a motorbike to convey the idea that the product is cool. This requires forming common associations that humans are aware of. Ye et al. \cite{Ye2017ADVISESA} devise the first attempts at this by learning associations between objects and their ideas. This is something that can currently be done by using Foundation models. There are other crucial visual metaphor datasets in the image domain, too. Benchmarks unifying different datasets for figurative language spanning metaphors, similes, idioms, etc., reveal VLMs' poor performance in contrast to literal tasks \cite{Saakyan2024UnderstandingFM, Akula2022MetaCLUETC, Zhang2021MultiMETAM}. The core reason is hallucinations and unsound reasoning, a score of improvement.

To summarize, we have explored the difficulties associated with comprehending metaphors. This issue becomes increasingly complex when trying to interpret symbols within images and even more so when these symbols are combined with actions over time in videos, presenting a considerably tougher challenge. The scarcity of datasets in the video domain can be linked to the inadequate performance of Foundation models in these tasks. Future investigations should aim to create Foundation models that incorporate symbolic reasoning abilities and utilize explainable AI along with human-in-the-loop approaches to ensure that model interpretations are in line with human understanding.

\textit{\textbf{Humor, Sarcasm \& Satire:}} Humor is deeply rooted in language and actions; thus, understanding requires reasoning about human traits and deep cultural context awareness. Here, we discuss some multimodal datasets dealing with this and also try to uncover some capabilities unlocked from Foundation models.

One of the first works to understand visual humor from abstract images comes from Chandrasekaran et al. \cite{Chandrasekaran2015WeAH}, who propose two novel datasets for this with incongruity theory (violation of expectations) \cite{Mihalcea2007TheMF} as a framework. Castro et al. \cite{Castro2019TowardsMS} was the first to propose the problem of sarcasm detection in a multimodal setting using videos, addressing the critical limitation of previous datasets that only relied on text-based analysis, ignoring important visual and audio cues in the process. They create a dataset (MUStARD). The visual features were pooled frame-wise using ResNet, and the text was processed using BERT. Temporal context and visual modality (facial expressions and gestures) are shown to be important through the baseline experiments. Along similar directions, Hasan et al. \cite{Hasan2019URFUNNYAM} introduced the dataset (UR-FUNNY), automatically understanding humor in a multimodal setting. The dataset is constructed from TED talks with annotations using text and audio, along with facial expressions. Glove word embeddings are used along with facial action units for feature extraction, which are then fused and fed to LSTMs. Baseline experiments demonstrate the importance of using all the modalities along with the understanding of context. Patro et al. \cite{Patro2021MultimodalHD} propose a benchmark to understand humor from sitcom videos, combining video and text-based analysis. BERT with C3D (for visual cues like facial expression and scene dynamics) is used for feature extraction, which is then fused to make the prediction. The authors do highlight the failure cases of the baseline model, which lacks the understanding of cultural context (missing the references to `Slumdog Millionaire') and also visual outliers (funny costumes) as humor situations. Foundation models may be added as an external knowledge base and can be leveraged for these. 
Kumar et al. \cite{Kumar2022WhenDY} present a sarcasm detection dataset that aims to detect sarcastic conversations and give explanations. Audio features are essential for capturing tone alongside visual features like gestures and facial expressions. ResNext-101 and openSMILE are used for feature extraction and fused with a decoder model like BART for reasoning. The model suffers from identifying the targets (who is being mocked), motivating a better deep dive into the problem, as it requires a deeper understanding of visual context and retention. Ko et al. \cite{Ko2023CanLM} introduce a dataset for humor detection in user-generated content covering a wide range of scenarios, instead of special scenarios like sitcoms, stage talks, etc. Foundation models are leveraged for this task. The authors convert all modalities to text via different captioning methods, which LLMs then reason over. Thus, when audio is converted to tags, it loses not only temporal context but also a significant amount of nuance in communication and humor. Failure cases include object/activity misidentification, leading to bad guesses. Thus, research in LLMs to understand objects correctly is vital for such scenarios. Recent datasets and benchmarks test the ability of Foundation models to comprehend humor and satire for a sequence of images \cite{Hu2024CrackingTC, Nandy2024YesButAH}. LLMs with rich descriptions perform better than VLMs. This could be attributed to the information loss due to the modality gap.

We discussed various methods to address the automatic understanding of humor and related figures. While models have evolved, their failure cases highlight significant limitations in grasping cultural references and identifying sarcasm targets. The reliance of Foundation models on text for reasoning creates a bottleneck and a \textit{modality gap}. To bridge this gap, further progress must address this limitation of Foundation models and reason over all modalities. Extensive pre-training to capture cultural nuances is also an essential step in this direction. Future research must also address the subjective nature of humor and reasoning when it becomes toxic. Developments in this direction can help filter out harmful content from content-sharing websites.

\subsection{Persuasion}
The art of persuasion revolves around campaigning, whether in politics or commerce. It involves real-time communication that emphasizes dialogue and utilizes pre-produced videos along with visual media. In this context, we explore the evolution of various persuasion strategies within a multimodal framework. Visual advertisements occasionally incorporate unconventional objects to catch viewers' attention (like an owl made of coffee), which are not typically seen but effectively communicate the commercial message \cite{Hussain_2017_CVPR}. The opening Figure \ref{fig:abstract_concepts} shows an example of such an atypical object. The first works in this direction come from Ye et al. \cite{Ye2017ADVISESA}, who decode symbolism in image advertisements. Guo et al. \cite{Guo2021DetectingPA} advance understanding of atypical objects, framing them differently from anomaly detection. The authors design a relative spatial transformer architecture to capture object-context compatibility and spatial interactions, therefore showing improvements in understanding atypical objects. While comprehensive studies in the video domain are lacking, we will highlight existing studies in the image domain here. We hope that future work also addresses this in the video domain. In \cite{liu2019generating}, the authors propose a method to generate persuasive storylines for video ads using the Wundt curve \cite{berlyne1960conflict} to model persuasiveness. They quantify information (via structural dissimilarity \cite{Wang2004ImageQA}), attractiveness (using NIMA \cite{Talebi2017NIMANI}), and emotion (arousal via MobileNet \cite{Howard2017MobileNetsEC} trained on \cite{Kim2017BuildingEM}). Instead of processing video directly, image-based models extract these features to learn a simple function based on the Wundt curve. This approach shows how abstract concepts can be modeled and transferred from images to video, even in a low-data regime.

Images and videos shared on different media channels are vital to political campaigning. Previous studies also link facial cues to politicians' personality traits and election outcomes, such as \textit{attractive}, \textit{competent}, etc. \cite{Joo2015AutomatedFT}. Political images are shared on the internet with communicative intent to demonstrate leaders' capabilities and traits (\textit{trustworthiness, energetic}, etc.) and involve persuasive strategies to build trust and following. The earliest work in this direction involving images comes from Joo et al. \cite{Joo2014VisualPI}, who propose understanding this using different visual cues like facial features, gestures, and scene context. Body gestures, when included, showed further improvement in the scores, as do the developments in deep learning as visual feature extractors \cite{Huang2016InferringVP}. Such research can be easily plugged into video domains and pave the way for understanding leaders' persuasiveness and effectively predicting the election outcome.

Real-life political campaigning also opens avenues for discovery in this direction. Among the first works to understand the political influence are those by Siddiquie et al. \cite{Siddiquie2015ExploitingMA}, who explore the persuasiveness of real-world political campaigns through videos (Rallying a Crowd \cite{chisholm2015audio}). Feature extractors for different modalities, detecting the type of speech (\textit{calm, agitated} and crowd reactions using spectrogram, CNN trained on ImageNet, and visual sentiment ontology were used for visual features.
In contrast, sentiment signals were taken from YouTube comments. Analysis showed that audio modality is the prime indicator, while visual modality helps further determine the impact.
Bai et al. introduce the problem of debate outcome prediction with indicators of persuasiveness (Intensity of Persuasion Prediction) in the pipeline \cite{Bai2020M2P2MP}. This shows yet again how monitoring persuasiveness can help in different domains, such as debates and, by extension, political debates. Recently, Liu et al. \cite{liu2025paladin} propose a taxonomy and dataset for understanding communication techniques used in political advertisements via classifying and segmenting the video clips. These include orating slogans, changing colors to evoke emotion, and other rhetorical strategies. The videos are chosen such that videos with purely verbal rhetoric with no visual manifestation are not included. Since these are highly abstract, the performance is poor for this benchmark, close to a random baseline. Similar visual features but different symbolic imagery (\textit{Implication of Emotion \& Emotion Mirroring}) are hard cases for models to detect. Also while the models are good at detecting cinematic styles (colors, rapid-cuts) the overarching persuasive technique is still hard to understand. The authors highlight the need to understand subjectivity and understanding deeper semantics of video.

Various taxonomies of persuasion strategies are presented in different domains, such as in image advertisements \cite{Singla2022PersuasionSI, Hussain_2017_CVPR}, political campaigns \cite{Liu2022ImageArgAM}, and memes \cite{Kumari2023ThePM}. Such taxonomies can help Foundation models predict persuasion strategies in zero-shot classification settings, as demonstrated in \cite{Bhattacharya2023AVI}. Future research could also lead to developing culturally-aware Foundation models capable of recognizing and interpreting culturally specific persuasive strategies.

\begin{figure}
    \centering
    \includegraphics[width=\linewidth]{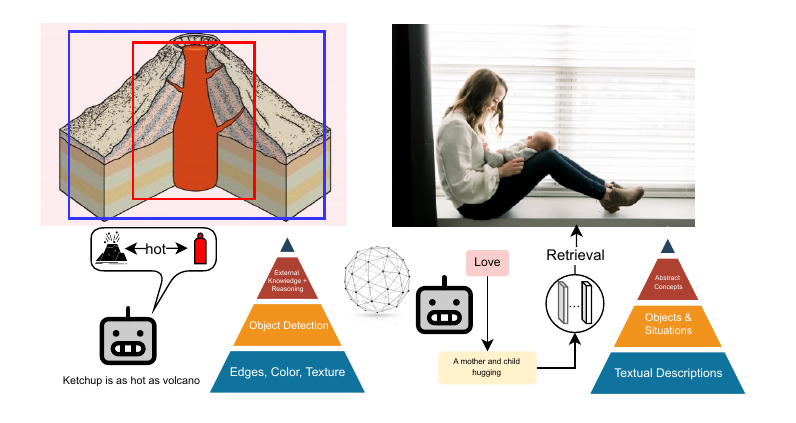}
    \caption{Example images used for visual metaphor detection and abstract concept image-retrieval from \cite{Akula2022MetaCLUETC} and \cite{Cerini2024RepresentingAC}: Foundation models help bridge the semantic gap between low- and high-level descriptions using context and external knowledge, e.g., ``Ketchup is as hot as a volcano" (left) and ``Love." (right).}
    \label{fig:enter-label}
\end{figure}

\setcounter{table}{1}
\begin{table*}[t]
\centering
\renewcommand{\arraystretch}{1.25} 
\Large 
\resizebox{\textwidth}{!}{%
\begin{tabular}{@{}lllllllr r r@{}}
\toprule
\textbf{Benchmark} &
\textbf{\# Videos} &
\textbf{Len (s)} &
\textbf{\# Ann.} &
\textbf{Task} &
\textbf{Domain} &
\textbf{Metric} &
\makecell[r]{\textbf{Performance:}\\\textbf{Closed-Source}} &
\makecell[r]{\textbf{Performance:}\\\textbf{Open-Source}} &
\makecell[r]{\textbf{Performance:}\\\textbf{Humans}} \\ \midrule

AdsQA \cite{long2025adsqa} &
1544 &
52.9s &
7859 &
QA &
\begin{tabular}[c]{@{}l@{}}Concepts, Emotions, Semantic Themes,\\ Persuasion, Audience Modeling \end{tabular} &
Accuracy &
\begin{tabular}[r]{@{}r@{}}GPT-4o (29.4) \\ Gemini 2.5 Pro (35.9)\end{tabular} &
\begin{tabular}[r]{@{}r@{}}Qwen2-VL-7B (17.2)\\ Qwen2.5-VL-7B (23.0)\\ Qwen2.5-VL-72B (34.7)\end{tabular} &
51.3 \\
\midrule

VideoAds \cite{zhang2025videoads}&
200 &
79.6s &
1,100 &
QA &
\begin{tabular}[c]{@{}l@{}}Visual finding, summary and reasoning\end{tabular} &
Accuracy &
\begin{tabular}[r]{@{}r@{}}GPT-4o-text (21.27)\\ GPT-4o (66.82)\\ Gemini 2.5 Pro (80.04)\end{tabular} &
\begin{tabular}[r]{@{}r@{}}Qwen2.5-VL-7B (54.69)\\ Qwen2.5-VL-72B (73.35)\end{tabular} &
94.27 \\

\midrule

BlackSwanSuite \cite{chinchure2025black} &
1655 &
8.83s &
15,469 &
QA &
Unexpected/surprising events &
Accuracy &
\begin{tabular}[r]{@{}r@{}}GPT-4o (65.1);\\ Gemini 1.5 Pro (58.7);\end{tabular} &
LLaVA-Video (55.9) &
90 \\
\midrule
Q-Bench-Video \cite{zhang2025q} &
1800 &
10–16s &
2378 &
QA &
\begin{tabular}[c]{@{}l@{}}Technical qualities, Aesthetic distortions\end{tabular} &
Accuracy &
\begin{tabular}[r]{@{}r@{}}GPT-4o (58.7)\\ Gemini-1.5-Pro (58.63)\end{tabular} &
\begin{tabular}[r]{@{}r@{}}m-Plug-Owl3 (52.39)\\ LLaVA-OneVision (Qwen2-7B) (51.7)\end{tabular} &
81.56 \\ 
\midrule

FunQA \cite{Xie2023FunQATS}&
4365 &
19s &
311K &
QA &
Surprise: Humour, Creative, Magic &
Accuracy &
\begin{tabular}[r]{@{}r@{}}GPT-4V (39)\\ GPT-4 (23)\end{tabular} &
Otter (26) &
- \\

\midrule

VMC \cite{kalarani2024unveiling} &
705 &
54s &
2115 &
Captioning &
Metaphor Captioning &
BLEU-4 &
- &
Video-LLaVA (16.88) &
- \\

\midrule

EMER \cite{lian2025affectgpt} &
31,327 &
2–5s &
1972 &
Classification &
\begin{tabular}[c]{@{}l@{}}Fine-grained emotion recognition\end{tabular} &
F1-Score &
- &
\begin{tabular}[r]{@{}r@{}}Emotion-LLaMA (64.17)\\ AffectGPT (74.77)\end{tabular} &
- \\

\midrule
MovieGraphs \cite{wang2025synergizing} &
7,637 &
44.28s &
Multiple &
Classification &
Social Relation Recognition &
Accuracy &
\begin{tabular}[r]{@{}r@{}}GPT-4o+Vicuna-13B-v1.5 (0.468)\\ Gemini-2-Pro+GPT4 (0.599)\end{tabular} &
- &
- \\

\bottomrule
\end{tabular}%
}
\caption{The table presents statistics from recent benchmarks covering various pillars and compares the performance of various open (Qwen \cite{yang2024qwen2}, Video-LLaVA \cite{lin2024video}),  and closed (GPT-4o \cite{hurst2024gpt}, Gemini 2.5 Pro \cite{comanici2025gemini}) source models. Human performance is included where reported in the original studies. Although Foundation models demonstrate strong results, the table indicates that significant performance gaps persist and further improvement is necessary.}
\label{tab:performance_table}
\end{table*}

\subsection{Framing Analysis}
Framing theory in communication explores the strategic presentation of information to influence perception and interpretation \cite{Borah2011ConceptualII}. This directly impacts how opinions are conveyed, how political leanings are perceived, and how potential misinformation can be detected across different modalities. This section briefly examines the visual and multimodal understanding of opinions and misinformation. 
 
\textbf{\textit{Opinion:}} Addressing the limitations of prior research and a lack of focus on conversational persuasion in multimodal settings, Park et al. \cite{Park2014ComputationalAO} propose a dataset (POM corpus) from the domain of people giving movie reviews with annotations like confidence and credibility. Different features such as pitch, pause rates (audio), unigrams and bigrams from text and facial action units, gaze, and emotion are used to train an SVM to predict whether the review is persuasive. A similar analysis is used to indicate the winners of a debate \cite{Brilman2015AMP}. Future works improve performance using a deep neural network instead of SVM, showing improvement on the dataset \cite{Nojavanasghari2016DeepM}. Zadeh et al. \cite{Zadeh2016MOSIMC} introduce MOSI, the first multimodal dataset for fine-grained sentiment intensity and subjectivity analysis in online videos. This is essential in jointly understanding opinions and sentiments in video content. The authors used different features such as facial action units, head pose, emotional annotations, voice pitch, and conversational pauses to determine the sentiments.

Understanding political leanings is also an interesting problem closely related to understanding opinions. Due to the lack of datasets in the video domain, we discuss the literature dealing with this in the image and text domains, motivating further work in videos. This includes understanding political leanings from campaign events and more. Thomas et al. \cite{Thomas2019PredictingTP} are among the first to tackle visual political bias and opinions, leveraging both text and image. The proposed task aims to predict the bias of image-text pairs (left-leaning or right-leaning). The authors use a two-stage pipeline, which includes predicting the labels utilizing both text and image in the first stage and using only the image in the second. The authors highlight that the simple occurrence of particular objects is not enough, and the portrayal also plays an important role. This means that the models need to understand the objects and the setting of the image, which becomes vital to understanding nuanced messages. SnchezVillegas et al. \cite{SnchezVillegas2021AnalyzingOP} tackle the problem of understanding political ideology and sponsoring political advertisements from political poster campaigns. BERT and EfficientNet are used for combined representation. However, the pipeline struggles with understanding nuanced political rhetoric. An improvement point suggested by the authors is also to discover ways to incorporate external knowledge as context, where Foundation models can help vastly.

\textbf{\textit{Misinformation:}} Predicting if the content is misinformation or a conspiracy is also an important theme to capture in political discourse. Recent works also derive inspiration from Framing Theory to detect such misinformation \cite{Wang2024DetectingMT}. With the rise of short video content platforms, it is essential, more so than ever, to combat misinformation in time before it can cause any damage.  Choi et al. \cite{Choi2021UsingTM} use topic modeling techniques to measure the difference between topics of titles (provider’s intent) and comments (audience reaction), a shift between topics being a good indicator of fake news. Visual feature extractors are used, but the performance is secondary to text modality. Further work can employ captioning models in this approach to improve the scores. Qi et al. \cite{Qi2022FakeSVAM} propose a dataset (FakeSV) in this direction and a pipeline to address this, including text, audio, video, user comments, and publisher profiles. The authors highlight that social context in the form of user profiles and comments (user behavior modeling) provides a valuable signal that improves the baseline scores, as shown by ablations. Feature extractors such as VGG, C3D, and BERT were used, and modality fusion by transformers was finally applied to detect if the news was fake. However, the authors highlight failure cases where misleading text on natural videos may lead to a missed case, such as labeling a police drill as a real-life incident, which aims to help us understand the intent of the actions of the actors in the shot. This analysis is missing from the paper and opens up the scope for future work. Sung et al. \cite{Sung2023NotAF} explore a similar direction to understand misleading news headlines. This is important as headlines sometimes exaggerate the content, which may not be present in the video. Therefore, understanding the headline and the visual modality in video is essential. Besides using feature extractors for video and text, authors also provide and use rationales as an input signal. Since the dataset has rationales as an annotation, future works may try to emulate them, similar to \cite{Singh2024TeachingHB}, to reason and improve the scoring of why a video headline is misleading.

Recent works like SNIFFER \cite{Qi2024SnifferML} aim to detect misinformation by verifying image-text pairs and also retrieving evidence while providing accurate judgments with explanations. The authors use vision transformers and ViT with a query transformer to bridge the gap between modalities. The pipeline checks for consistency between the text and image using the Google Vision API for entity verification. Webpages, where similar images were published, are retrieved and checked against the title, where the LLM makes a judgment. This is a nice idea of Foundation models as agents and opens the door for reasoning agents to detect misinformation.

\section{Discussion}
In the previous section of the paper, we explored various efforts to recognize and detect abstract concepts. Recognizing such abstract concepts requires models to emulate social intelligence, possess common-sense reasoning and human-level perception, and resolve ambiguities by providing an extensive knowledge base. Modeling efforts in this direction have evolved from using hand-crafted features to employing deep learning for automatic feature extraction and, more recently, leveraging large-scale pre-trained Foundation models with rich knowledge bases. The rise of Vision-Language Models and Multimodal Large Language Models has narrowed the gap between visual and linguistic modalities, enabling progress across various tasks, including classification, captioning, cross-modal retrieval, and content generation. This, along with the agentic capabilities of such models, opens new avenues for applications. We also found that many of these challenges are interrelated; for instance, identifying persuasive strategies or narrative elements often involves interpreting emotional cues, and at the same time, distinguishing pranks from fails requires understanding action intent. These observations highlight the need for a more robust foundation model capable of comprehending such abstract concepts. Research like \cite{Zhang2018EqualBN}, which focuses on understanding symbolism in advertisements, highlights the importance of integrating multiple abstract concepts and tasks, such as topic understanding, memorability, surprise, and low-level features like HOG, to bridge the semantic gap. Emerging comprehensive benchmarks like MV-Bench and MM-Bench \cite{Li2023MVBenchAC, Fang2024MMBenchVideoAL} further emphasize the necessity of Foundation models that can reason across varying levels of abstraction. Efforts to unify diverse perception-based datasets within a single framework reflect the hierarchical nature of human perception, from low-level features to high-level reasoning, and illustrate the interconnection among these concepts \cite{Fan2016APF}. This helps in bringing different datasets into a unified benchmark. We advocate for strong research in this direction, which brings automatic video understanding close to human-level intelligence.


\begin{figure*}[t!]
    \centering
    \includegraphics[width=\linewidth]{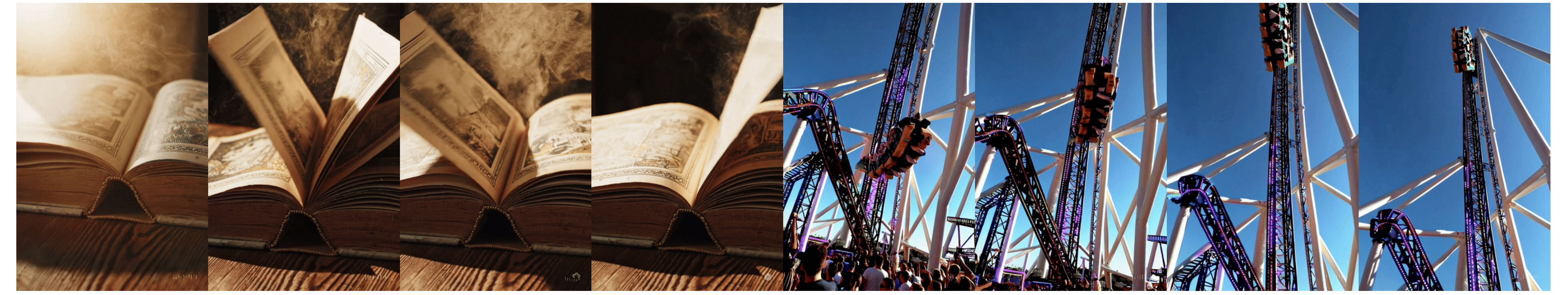}
    \caption{Video generation examples for the following prompts: \textit{Don’t judge the book by its cover} and \textit{Life is a roller coaster}, using SORA \cite{openai_sora_system_card_2024}. These examples illustrate how models can become stuck in literal interpretation rather than their abstract meaning.}
    \label{fig:generation}
\end{figure*}

\section{Challenges \& Future Directions} 
\textbf{\textit{Perception versus Reasoning}}: The relative order of complexity of these two tasks is inconclusive, with research supporting both sides \cite{wu2025friendsqa, ma2025iv}. We, however, argue that correct perception is vital for reasoning and, therefore, understanding abstract concepts. An example is to understand the correct objects in visual metaphor understanding from advertisements. Wrongly identified objects may lead to a completely different meaning. We therefore highlight both tasks as being equally important, as correct understanding relies on accurate reasoning, which in turn depends on perception.

\textbf{\textit{Subjectivity}}: Classification-based datasets employ majority voting to finalize labels, which often ignores the subjectivity of concepts, such as emotion, humor, and intent. Mechanisms to encode subjectivity and adapt training/evaluation paradigms to it are still scarce. Some steps in this direction include training models with annotator context (such as race and gender) \cite{anand2024don, fleisig2023majority} or flagging subjective attributes and classifying them beforehand \cite{homayounirad2025will}. The goal is to develop models that emulate how a diverse group of humans would respond, rather than collapsing to a single output, treating subjectivity as a feature rather than noise. Future research in this direction will directly benefit understanding abstract concepts.

\textbf{\textit{Metrics}}: As prediction outputs shift from classification to open-ended queries, the community needs to develop metrics that capture the correct semantic meaning and subjectivity beyond n-gram measures. Automated evaluations using LLM judges have been shown to yield different scores even when the input is the same \cite{haldar2025rating, Xie2023FunQATS}, indicating that this may not be a robust metric. Future research directions should focus on developing metrics that incorporate adaptive penalties, accommodate multiple subjective outputs, and extract the high-level semantics of two text outputs with ease.

\textbf{\textit{Cultural Bias}}: Recognizing abstract concepts directly benefits from understanding cultural context \cite{borghi2022abstract}. Vision Language models have been shown to demonstrate Western cultural bias \cite{ananthramsee} and an inability to incorporate multiple cultural or historical perspectives \cite{wang2025artragretrievalaugmentedgenerationstructured}. It is important to have models trained on culturally diverse and grounded datasets. However, such studies are more common in image space. Future directions can investigate such biases in video models and develop methods to mitigate them.

\textbf{\textit{Modality Gap}}: Many datasets for recognising abstract concepts are designed as multimodal from the outset to efficiently capture a range of signals from all modalities, a good example is the analysis of social media posts that combine text with images or videos (MINE \cite{yang2025uncertain}), as well as humor, sarcasm, and relationships containing narrations or background music (e.g. UR-FUNNY \cite{Hasan2019URFUNNYAM}, MUStARD \cite{Castro2019TowardsMS}, PERR \cite{gao2021pairwise}). Incorporating and interpreting multimodal signals has demonstrated the advantages; for instance, actions and gestures can convey intent in non-verbal scenarios \cite{Zhang2024MIntRec20AL}. However, due to extensive pre-training that may lead models to memorize datasets (as observed in NExT-QA \cite{Xiao2021NExTQANP} performance) or as a result of training artifacts, models often exploit shortcuts that bypass the vision modality altogether to achieve high benchmark scores (DeSIQ \cite{Guo2023DeSIQTA}). This highlights the need for model architectures that are firmly grounded in vision and capable of interpreting fine-grained cues such as facial expressions, body language, and advertisement strategies beyond company names to reason over videos and abstract concepts. Some works have also demonstrated the advantage of offloading the processing of non-textual modalities to other modules and using LLMs for reasoning on emotions (AffectGPT \cite{lian2025affectgpt}), which can be generalized for other concepts as well. Furthermore, there is a need to develop models and pipelines that integrate vision, language, audio, and knowledge (e.g., world knowledge captured in LLMs or domain knowledge encoded in knowledge graphs) enabling the extraction and synergistic integration of relevant information from each modality.

\textbf{\textit{Hallucinations}}: Foundation models are susceptible to hallucinations and often fail to recognize intended entities at their respective time stamps \cite{liu2024survey}. Accurate recognition of abstract concepts depends on the correct identification of lower-level semantics such as objects and scenes. Incorrect recognition of objects and scenes may result in models developing flawed reasoning, which can subsequently cause misidentification of the intended abstract concept. This issue is prevalent in both perception and reasoning, complicating error analysis because the models frequently respond with high confidence. Moreover, abstract concepts are more difficult to comprehend and are more susceptible to hallucinations, as their meanings are implicit, and responses may be influenced by subjective interpretation. Achieving performance beyond standard benchmarks requires models to be robust and adaptable to diverse scenarios while ensuring the reliability and reproducibility of results. Therefore, minimizing hallucinations in Foundation models is essential for reducing error propagation.

\textbf{\textit{Reasoning beyond the Chain-of-thought}}: Although the Chain-of-thought paradigm benefits many tasks, it seldom improves the recognition of abstract concepts and may even hinder performance \cite{chinchure2025black, long2025adsqa} This limitation arises because abstract concepts do not necessarily adhere to a sequential logic, such as if A then B, which is advantageous in other reasoning tasks like mathematics \cite{long2025adsqa}. Additionally, error propagation from the initial step can result in incorrect final predictions \cite{chinchure2025black} A promising direction involves reasoning over these concepts directly within a latent space. Furthermore, a query-conditioned Foundation model (deriving inspiration from Mixture of Experts \cite{mu2025comprehensive}) could be developed to unify all pillars of abstract concepts within its architecture, selecting relevant modules based on the query, such as separate modules for perception, emotions, and long-term memory.

\textbf{\textit{Temporal Grounding and Frame Sampling}}: Adjacent video frames are generally similar, while longer temporal contexts contain greater diversity in semantic information. Uniform sampling can hinder prediction accuracy by omitting fine details present in unsampled frames, whereas dense sampling significantly increases computational costs. Temporal information is particularly important for tasks involving rapid scene changes or complex visual phenomena, where inadequate temporal modeling leads to poor benchmark performance. Benchmarks focused on long video understanding have further driven research in this area. Recent advances in token compression methods such as memory-based approaches \cite{he2024ma} have demonstrated promise for capturing temporally fine-grained information in long videos. Many abstract concepts, including visual metaphors in advertisements or narrative progression in films, are inherently temporal. Although models may rely on spatial context as a shortcut, temporally grounded models capable of processing long videos, selecting relevant input tokens for reasoning, will be essential for improving abstract concept understanding, as this fundamentally depends on accurately modeling event sequences.

\textbf{\textit{Data Leakage}}: While it has been demonstrated that scaling the dataset leads to a performance boost in Foundation models \cite{kaplan2020scaling}, this raises the issue of dataset contamination, making it challenging to assess the true understanding capabilities of Foundation models. Multiple studies have highlighted the issue of data leakage and memorization in Foundation models \cite{ramos2025data}. This leads to inflated metrics and questionable performance of Foundation models, making it hard to verify the generalization capabilities. For example, FunQA \cite{Xie2023FunQATS} demonstrated that even without vision as a modality, GPT-4 performed at par with VLMs on NeXT-QA \cite{Xiao2021NExTQANP}, indicating a possible form of memorization. A similar case is also highlighted by Ghermi et al. \cite{Ghermi2024LongSS}, which showed that movie titles were sufficient to achieve high accuracy on benchmarks like LVU \cite{Wu2021TowardsLV} and MovieQA \cite{Tapaswi2015MovieQAUS}. This issue may slow down the development of models designed to understand abstract concepts, as it can create a false impression of advanced comprehension capabilities. Since understanding abstract concepts is considerably more challenging, high benchmark scores achieved primarily through memorization may indicate minimal genuine progress. Some approaches attempt to address this issue by shuffling color channels \cite{lu2024clean} or randomizing captions during training \cite{jayaraman2024deja} to prevent memorization; however, this remains an active area of research. There is a need to develop robust benchmarks and training protocols to enhance generalization and avoid memorization. It is also vital to diagnose and develop appropriate evaluation protocols for cases involving data leaks.

\begin{figure}[t]
    \centering
    \includegraphics[width=\linewidth]{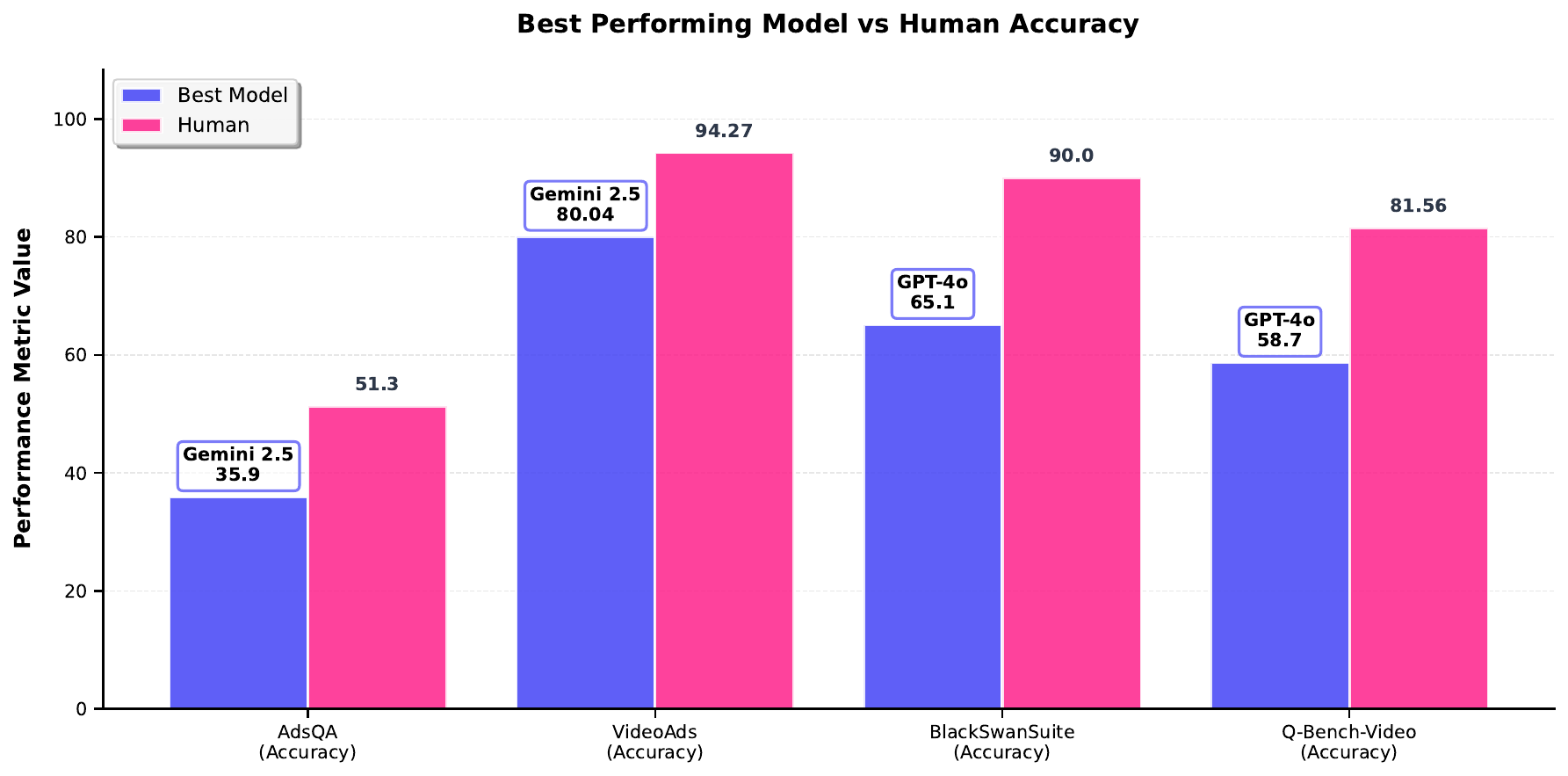}
    \caption{This figure illustrates the performance gaps between the best-performing models, including both open and closed source, and human accuracy when using Foundation models. The observed gap emphasizes significant potential for improvement. The figure also highlights the zero-shot performance of Foundation models. Pursuing the future directions outlined in the relevant section may help to reduce this gap.}
    \label{fig:performance_figure}
\end{figure}

We have observed a significant shift in the ability to understand and process abstract concepts, driven by substantial advancements in benchmarks and models. These concepts, initially treated as labels, can now be addressed through open-ended questions and tasks that evolve from simple classification to a broad spectrum of captioning, retrieval, question-answering, and more. Yet, the performance highlighted in the Figure \ref{fig:performance_figure} demonstrates that challenges remain unresolved, which are integral to the ongoing effort to understand abstract concepts and, more broadly, video understanding. Foundation models, due to their large-scale pretraining, provide substantial context for reasoning over open-ended queries. This advancement enables the development of benchmarks that are more complex than standard action recognition. By following the Foundation model training paradigm, it is possible to develop a model capable of understanding abstract concepts, accounting for their subjectivity, and serving as a comprehensive processing unit for reasoning across multiple dimensions of abstraction. This is the next frontier in advancing automatic understanding of videos.

As new benchmarks are developed, solving them increasingly requires reasoning at the level of multiple types of abstract concepts. Annotations for aesthetics, emotions, cinematic styles, and underlying messages are sometimes grouped within a single benchmark (MOGaze \cite{tores2025mobygaze}, Paladin \cite{liu2025paladin}). Although each annotation may appear to represent a distinct task, many co-occur and share similar abstract characteristics. For example, a video's climax typically elicits stronger emotions, while the use of grayscale shots can imply poor performance by political candidates. Tasks involving abstract ideas often share common elements, such as humor or plot twists in advertisements, or causal reasoning that advances a narrative. Because Foundation models frequently encounter open-ended questions, they must be capable of understanding multiple, if not all levels of abstraction. Unifying all tasks under one umbrella would lead to a better pathway for human-like reasoning. This requires developing benchmarks with rich fine-grained annotations, Foundation models that can reason over multiple levels of abstraction, and evaluation metrics that account for subjectivity and different contexts, such as cultural and historical.

Video Generation has also experienced significant progress in recent years, and generating videos based on abstract keywords is an interesting direction to explore as well. LLMs have demonstrated their abilities to bridge the high-level textual semantics and low-level visual concepts for text-to-image generation \cite{liao2024text, chakrabarty2023spy, fan2024prompt}. A common recurring challenge for Foundation models is the attribute-object binding \cite{chakrabarty2023spy, ramesh2022hierarchical}, which associates attributes with the correct objects. Some researchers \cite{Liao2023TexttoImageGF} attempt to address this issue by drawing parallels to artistic generation, utilizing WordNet \cite{Miller1995WordNetAL} as an additional bridge. More recently, large-scale datasets containing videos with abstract captions \cite{xieabstext2video} have been proposed. Works like this will pave the way for future models to directly understand the text queries and generate videos based on abstract concept themes. We also qualitatively highlight a failure case in Figure \ref{fig:generation}. However, a detailed discussion on video generation is beyond the scope of this survey.

\section{Conclusion}
In this paper, we explored a range of abstract concepts and their occurrence within the video and multimodal domain. By integrating both automated and manual literature survey methods, we developed a comprehensive taxonomy that encapsulates a wide array of relevant topics. Our analysis reveals recurring patterns that reflect the research community’s ongoing efforts to close multiple conceptual gaps, such as the semantic and affective gap, while also addressing gaps related to intent, modalities, commonsense reasoning, social understanding, and narrative structure through diverse methodologies. Foundation models emerge as a promising avenue in this pursuit. With the advent of new benchmarks, we advocate for the development of more robust Foundation models capable of recognizing and interpreting abstract concepts across multiple semantic levels. Such advancements, we argue, are essential for narrowing the gap between low-level signal processing approaches and models that aspire to achieve human-like understanding.

\section*{Acknowledgments}
This work was supported by the UvA Data Science Centre, as part of the HAVA Lab.

\section*{Data Availability}
All data used in this survey were collected using the Semantic Scholar API, an open data platform featuring a large database of scholarly data.



\onecolumn
\begin{appendices}
\section{List of Conferences}\label{secA1}
The list of conferences used for this survey is listed here in Table \ref{conference_table}. We focus on the conferences having \textit{A$^*$, A} or \textit{B} rating according to International CORE Conference Rankings (ICORE).
\begin{table*}[h]
\caption{List of Conferences used for filtering papers from the Semantic Scholar Database}\label{conference_table}%
\begin{tabular}{@{}lc@{}}
\toprule
Conference Name &  Core Rating  \\
\midrule
AAAI Conference on Artificial Intelligence (AAAI) & A*  \\
Advances in Neural Information Processing Systems (NeurIPS) & A* \\
International Conference on Computer Graphics and Interactive Techniques (SIGGRAPH) & A* \\
International Conference on Learning Representations (ICLR) & A* \\
International Joint Conference on Artificial Intelligence (IJCAI) & A* \\
European Conference on Computer Vision (ECCV)& A*\\
Computer Vision and Pattern Recognition (and workshops) (CVPR) & A* \\
ACM Multimedia (ACMMM) & A* \\
The Web Conference (WWW) & A* \\
Knowledge Discovery and Data Mining (KDD) & A*\\
ACM SIGIR Conference on Research and Development in Information Retrieval (SIGIR) & A* \\
Conference on Empirical Methods in Natural Language Processing (EMNLP)  & A* \\
International Conference on Computer Vision (ECCV) & A* \\
International Conference on Machine Learning (ICML) & A* \\
Association for Computational Linguistics (ACL) & A* \\
British Machine Vision Conference (BMVC) & A \\
International Conference on Information and Knowledge Management(CIKM) & A \\\
IEEE Workshop/Winter Conference on Applications of Computer Vision (WACV) & A \\
Conference of the European Chapter of the Association for Computational Linguistics (EACL)& A \\
European Conference on Artificial Intelligence (ECAI) & A \\
European Conference on Information Retrieval (ECIR) & A \\
International Conference on Artificial Intelligence and Statistics (AISTATS) & A\\
North American Chapter of the Association for Computational Linguistics (NAACL) & A \\
IEEE International Conference on Multimedia and Expo (ICME) & A \\
ACM Conference on Recommender Systems(RecSys) & A\\
Web Search and Data Mining (WSDM) & A \\
International Conference on Multimedia Retrieval (ICMR) & B \\

\botrule
\end{tabular}
\end{table*}
\end{appendices}

\end{document}